\documentclass[a4paper,fleqn]{cas-sc}

\usepackage[authoryear,longnamesfirst]{natbib}

\usepackage{booktabs}
\usepackage{multirow}
\usepackage{graphicx}  
\usepackage{array}     

\def\tsc#1{\csdef{#1}{\textsc{\lowercase{#1}}\xspace}}
\tsc{WGM}
\tsc{QE}

\begin{document}
\let\WriteBookmarks\relax
\def\floatpagepagefraction{1}
\def\textpagefraction{.001}
   
\shorttitle{}   

\shortauthors{Xiaolin Gou et~al.}


\title[mode = title]{An Effective UNet Using Feature Interaction and Fusion for Organ Segmentation in Medical Image}

\tnotemark[1]
\author[1, 2]{Xiaolin Gou}
\ead{2022226045015@stu.scu.edu.cn}
\affiliation[1]{organization={National Key Laboratory of Fundamental Science on Synthetic Vision, College of Computer Science, Sichuan University},
    city={Chengdu},
    postcode={610000}, 
    country={China}}
\affiliation[2]{organization={College of Computer Science, Sichuan University},
    city={Chengdu},
    postcode={610000}, 
    country={China}}
    
\author[1, 2]{Chuanlin Liao}[style=chinese]
\ead{chuanlinliao@stu.scu.edu.cn}

\author[2]{Jizhe Zhou}[style=chinese]
\ead{jzzhou@scu.edu.cn}

\author[3]{Fengshuo Ye}[style=chinese]
\ead{2022141080058@stu.scu.edu.cn}
\affiliation[3]{organization={College of Software Engineering, Sichuan University},
    city={Chengdu},
    postcode={610000}, 
    country={China}}
\author
[1, 2]
{Yi Lin}
\cormark[1]
\ead{yilin@scu.edu.cn}

\cortext[cor1]{Corresponding author}
\tnotemark[1,2]

\begin{abstract}
Nowadays, pre-trained encoders are widely used in medical image segmentation due to their strong capability in extracting rich and generalized feature representations. However, existing methods often fail to fully leverage these features, limiting segmentation performance. 
In this work, a novel U-shaped model is proposed to address the above issue, including three plug-and-play modules. 
A channel spatial interaction module is introduced to improve the quality of skip connection features by modeling inter-stage interactions between the encoder and decoder.
A channel attention-based module integrating squeeze-and-excitation mechanisms with convolutional layers is employed in the decoder blocks to strengthen the representation of critical features while suppressing irrelevant ones.
A multi-level fusion module is designed to aggregate multi-scale decoder features, improving spatial detail and consistency in the final prediction.
 Comprehensive experiments on the synapse multi-organ segmentation dataset and automated cardiac diagnosis challenge dataset demonstrate that the proposed model outperforms existing state-of-the-art methods, achieving the highest average Dice score of 86.05\% and 92.58\%, yielding improvements of 1.15\% and 0.26\%, respectively.
 In addition, the proposed model provides a balance between accuracy and computational complexity, with only 86.91 million parameters and 23.26 giga floating-point operations.
\end{abstract}


\maketitle

\section{Introduction}
 Organ segmentation in medical images plays a vital role in the field of healthcare, providing essential structural information for diagnosis, treatment planning, and clinical decision-making. Accurate delineation of organs is fundamental for downstream tasks, such as radiation therapy planning, surgical navigation, and quantitative assessment of organ morphology and function. Traditionally, organ segmentation has been performed manually by expert clinicians, which is time-consuming and subject to variability due to human error and inter-observer differences.
In contrast, automatic organ segmentation methods can substantially reduce the time and effort required for image analysis, thereby improving clinical efficiency. Moreover, by minimizing human-related subjectivity, these methods offer enhanced consistency, reproducibility, and accuracy, which are especially critical in high-stakes clinical environments.

In recent years, deep learning has advanced across various fields. Convolutional neural networks (CNNs) can achieve feature extraction and representation for images, thus eliminating the requirements for handcrafted features. 
In this context, CNN-based automatic segmentation tools implement image segmentation by learning image features from large amounts of training samples through neural architecture, which can be generalized to new tasks with considerable high performance (\cite{Azadetal2022}). 
UNet (\cite{Ronnebergeretal2015}) became the most popular framework in medical image segmentation due to its simple yet effective architectural design and high performance, which can be applied to various modalities in medical images, including CT, MRI, X-ray, PET, etc. 
UNet is implemented by an encoder-decoder architecture with skip connections. The encoder gradually transforms the images into abstract representations by multi-level feature extraction and down-sampling operations.
The decoder predicts the segmentation masks based on the abstract representation, in which the up-sampling operations are leveraged to recover image resolution to generate pixel-wise masks.
As the core component of the UNet, the skip connection combines the features of the adjacent encoder stage and decoder stage to achieve high-efficiency learning.

Although the UNet models demonstrate the desired performance in medical image segmentation tasks, they still cannot capture global contextual information due to the limited receptive field.  To address this issue, the Transformer blocks (\cite{Vaswanietal2017}) are incorporated into UNet architectures to enhance the global feature integration and contextual understanding, such as SwinUNet (\cite{Caoetal2022}), TransUNet (\cite{Chenetal2021}), MISSFormer (\cite{Huangetal2021}), UNRTR (\cite{Hatamizadehetal2022}), and so on. However, the Transformer lacks inductive biases in CNNs, such as translation invariance and local feature learning ability, which makes it hard to achieve the expected performance with insufficient training samples.
 Recent advances in the pre-training Vision Transformer (ViT) (\cite{Dosovitskiyetal2020}) (e.g., Efficientnet (\cite{TanLe2019}), ConvNeXt (\cite{Liuetal2022}), DeepViT (\cite{Zhouetal2021}) have enabled the extraction of rich and generalizable visual features from large-scale datasets. These pre-trained encoders significantly reduce the reliance on task-specific labeled data and offer strong performance for downstream tasks. However, efficiently leveraging such powerful encoder features in the decoder remains a critical challenge, particularly for pixel-wise prediction tasks like medical image segmentation.

 To address this, we propose a novel U-shaped model (FIF-UNet) to fully utilize multi-level semantic features extracted by the encoder. The proposed model integrates three purpose-designed components to enhance the information flow between encoder and decoder in a complementary manner: (1) Channel Spatial Interaction (CSI) module, (2) Cascaded convolution-SE (CoSE) module, and (3) Multi-Level Fusion (MLF) module.

 The CSI module is designed to explicitly address the semantic gap between encoder and decoder features, cascading a channel interaction unit (CIU) and a spatial interaction unit (SIU) to refine skip connection features. This design enables the model to learn correlations between semantic representations at different levels, thereby delivering more accurate and informative feature maps to the decoder.
In the decoder path, the CoSE module is proposed to enhance feature selection by integrating channel attention through Squeeze-and-Excitation (SE) blocks (\cite{Huetal2018}) into convolution operations. The CoSE enables the model to adaptively emphasize task-relevant features while suppressing background noise, which is crucial for delineating complex anatomical structures.
In addition, to mitigate the detail degradation caused by upsampling operations, the MLF module is proposed to aggregate semantic features across decoder layers. Unlike standard progressive upsampling, MLF performs feature interactions across multiple scales to retain fine-grained information and contextual coherence in the predicted segmentation maps.

 In the proposed model, these three components form a coherent feature interaction and fusion strategy to enhance both the representational capacity and generalization ability. Extensive experiments on Synapse and ACDC datasets show that the proposed model outperforms selective competitive baselines, achieving average Dice scores of 86.05\% and 92.58\%, respectively. The ablation results further confirm the dedicated performance contributions of CSI, CoSE, and MLF.

 In summary, the main contributions of this work are shown as follows:

 (1) A new U-shaped model, FIF-UNet, is proposed to improve feature exploration from pre-trained encoders through explicit feature interaction and fusion strategies in the decoding process.

 (2) The CSI module is designed to enhance the quality of features passed through skip connection, thus providing the decoder with more accurate and semantically enriched feature maps.

 (3) The CoSE module is integrated into the decoder to enhance discriminative feature learning by combining convolutional operations with channel attention mechanisms.

 (4) The MLF module is introduced to efficiently aggregate multi-scale decoder features, improving the model's ability to recover detailed structures in segmentation outputs.

\section{Related Work}
As the core blocks of the UNet, CNN-based models were the dominant methods for various computer vision tasks. 
The Transformer-based models were also regarded as mainstream due to their recent advancements across many artificial intelligence tasks.
Integrating these modules with the UNet emerges as an enhanced strategy to improve the performance of medical image segmentation.
In this section, related works are organized as follows:
\subsection{CNN models}
Before the ViT model in 2020, CNN-based UNet models were the dominant approaches in the field of medical image segmentation, which efficiently capture local features through convolution operations. 
However, the original UNet suffered from limited feature extraction capability and the semantic gap between the encoder and decoder. 
To address these issues, 
the encoder or decoder modules of the UNet were improved to enhance feature learning. 
In DUNet (\cite{Jinetal2019}), the convolution layers of the original UNet were replaced by the deformable convolution layer to capture intricate features. 
The inception layers of the Google-Net (\cite{punnAgarwal2020}) were applied to automate the selection of the variety of layers in the deep network.
However, the mentioned improvements were mainly based on local convolution operations, with only a weak ability to capture global contexts.

Other works focused on adjusting the skip connections to alleviate the semantic gap between the encoder and decoder. 
The Group Aggregation Bridge module (GAB) in EGE-UNet (\cite{ruanetal2023}) effectively fused multi-scale information by grouping low-level features, high-level features, and a mask generated by the decoder at each stage.
The densely connected skip connections were designed to aggregate features of different semantic scales in UNet++ (\cite{Zhouetal2019}), resulting in forming a highly flexible feature fusion scheme. 
The Attention Gate (AG) was introduced to automatically focus on target structures with different shapes and sizes by employing a large receptive field and semantic contextual information in Attention U-Net (\cite{Oktayetal2018}).  
A two-round fusion module (i.e., top to bottom and bottom to top) in the skip connections was performed to reduce the semantic gap in FusionU-Net (\cite{Lieatl2024}).
 BRAU-NET++ (\cite{lan2024brau}) enhanced feature fusion by element-wise addition between encoder and decoder features, followed by channel and spatial attention. The proposed Skip Connection Channel-Spatial Attention (SCCSA) mechanism improves the model’s ability to highlight important features. However, most existing methods focus on enhancing encoder-side features or unidirectional fusion strategies, while overlooking the importance of explicitly modeling semantic correlations between encoder and decoder features for more effective integration of skip connections.

\subsection{Vision Transformer models}
Transformer was initially proposed for natural language processing and opened up new avenues for innovation in computer vision tasks (\cite{Dosovitskiyetal2020}).
The Transformer block allows each element in the input sequence to focus on all other elements by a self-attention mechanism, thus constructing pure Transformer models to effectively adapt to complex image scenes and objects of various sizes over CNNs.
For example, the hierarchical Swin Transformer with shifted windows was used as a base block to learn global and distant semantic interactions in SwinUNet (\cite{Caoetal2022}). 
Gating mechanisms were added to the axial-attention to learn relative positional coding, to further accurately encode long-range interactions in MedT (\cite{Valanarasuetal2021}).
In MISSFormer (\cite{Huangetal2021}), efficient self-attention and enhanced mix-FFN were introduced to construct an enhanced Transformer block for aligning features with higher consistency.
Inspired by dilated convolutions, a dilated Transformer was proposed to perform global self-attention in a dilated manner in D-Former (\cite{Wuetal2023}), which expanded the receptive field and reduced computational cost without adding patch blocks.
However, compared with CNNs, the pure Transformer model is limited by the learning of local features, which impacts the accurate capture of detailed features, especially in the delicate medical image segmentation task.
\subsection{Hybrid CNN-Transformer models}
The hybrid CNN-Transformer models utilize the advantages of the Transformer in capturing long-range dependencies and global information while retaining the efficacy of CNNs in handling local features. 
This unique combination enables the hybrid models to achieve cutting-edge performance in various tasks, especially in medical image segmentation. 
In TransUNet (\cite{Chenetal2021}), CNNs were employed to extract local features to project the output into labeled image blocks, which were then fed into a cascaded Transformer module to learn global features.
In TransBTS (\cite{Wangetal2021}), the Transformer was introduced at the bottleneck connection to model global contexts on local feature maps from the CNN encoder.
Considering the high computing cost of the Transformer, in MTU-Net (\cite{Wangetal2022}), the CNN operations were applied in upper layers to focus on local relations, while the Mixed Transformer module was designed in the deeper layers with smaller spatial dimensions.
In FCT (\cite{tragakisetal2023}), each stage of the UNet processed its input in two steps, i.e., extracting long-range semantic dependencies by Transformer blocks, and capturing semantic information across different scales using dilated convolutions with certain dilated ratios.
In TMU (\cite{Azadetal2022a}), the hierarchical local and global features were extracted by CNN and Transformer, which were fed into the contextual attention module to adaptively recalibrate the representation space to highlight the information regions.
Although combining CNN can improve the efficiency of feature extraction, the hybrid model still has high computational complexity, and it is challenging to properly integrate the advantages of CNN and Transformer.
\begin{figure*}[!h]
    \centering
    \includegraphics[width=\linewidth, keepaspectratio]{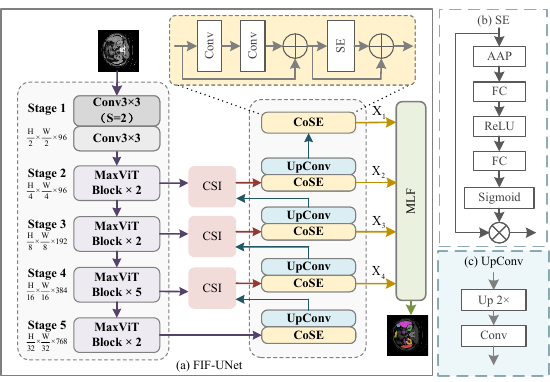}
    \caption{ (a) The architecture of FIF-UNet, (b) Squeeze-and-Excitation Network, (c) UpConv Module. “Conv” represents a standard combination of convolution, batch normalization, and activation.}
    \label{overall}
\end{figure*}

\section{Method}
\subsection{Overall architecture}

\begin{figure*}[!h]
    \centering
    \includegraphics[width=\linewidth, keepaspectratio]{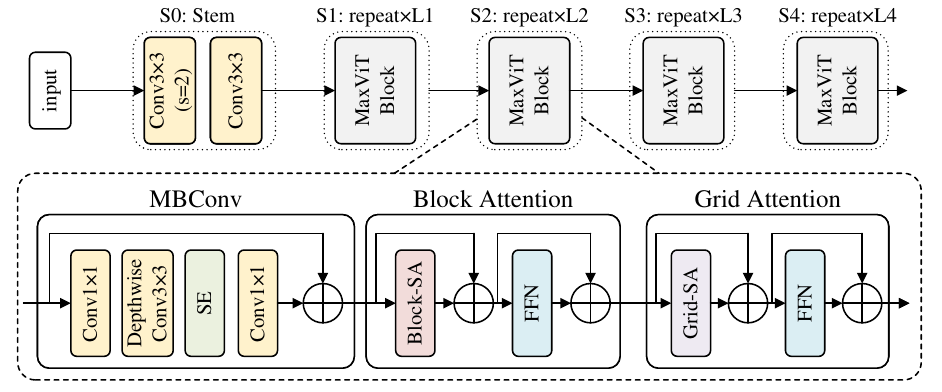}
    \caption{ MaxViT architecture. L1, L2, L3, and L4 represent the number of cascaded MaxViT blocks in the four stages, respectively.}
    \label{MaxViT}
\end{figure*}

 The overall architecture of the FIF-UNet is shown in Figure \ref{overall}(a), employing a U-shaped structure with symmetric encoder-decoder modules. 
To ensure completeness and reproducibility, a detailed layer-wise configuration is additionally provided in Table \ref{architecture}.
In the encoder, a hybrid CNN-Transformer model, called MaxVit-S (\cite{Tuetal2022}), serves as the backbone network, which is pre-trained on the ImageNet dataset utilizing an image classification task. 
Compared to full self-attention, the MaxViT is implemented based on blocked local and dilated global attention to capture both the local and global features, which can be calculated by only linear complexity ($O(n)$, $n$ is the spatial size of an input image). 
 As illustrated in Figure \ref{MaxViT}, the encoder network consists of 5 stages, including a stem stage and four cascaded MaxViT stages.
In the stem stage, two convolution layers are with 96 channels and a kernel size of 3. The stride of the first CNN layer is set to 2 to downsample the input image resolution.
The configurations of the MaxViT are with the $\left\{2, 2, 5, 2\right\}$ blocks and generate the feature maps with $\left\{96,192,384,768\right\}$ channels, respectively. 

In the skip connections, a CSI module is proposed to dynamically recalibrate the feature maps by the designed CIU and SIU, with the objective of obtaining the informative target features.
In the decoder network, each decoder stage is constructed based on the CoSE module and UpConv module. 
The proposed CoSE module aims to enhance the representation of critical features by incorporating the SENet mechanism into CNNs.  
The UpConv module upsamples the resolution of the CoSE outputs by bilinear interpolation, followed by a convolution layer to refine the up-sampled feature maps, as in Figure \ref{overall}(c). 
Instead of predicting the segmentation tasks based only on the last decoder stage, in this work, a MLF module is innovatively proposed to effectively fuse the outputs of the decoder stages to enhance the 
segmentation details by integrating intra- and inter-class features.

\begin{table}[h]
\renewcommand{\arraystretch}{1.2}
\centering
\caption{ Detailed architecture of FIF-UNet. Size: C × H × W(C: channels, H: height, W: width), Res: Residual operation, Bi: Bilinear, CC: Channel Concatenation.}
\resizebox{1\textwidth}{!}{
\begin{tabular}{lc|l|ll}
\toprule
\textbf{Network} & \textbf{Stage} & \textbf{Operation} & \textbf{Input Size} & \textbf{Output Size}  \\
\midrule
\multirow{5}{*}{Encoder} & 1 & Conv3×3(stride=2) + Conv3×3(stride=1) & 3 × 256 × 256 & 96 × 128 × 128 \\
& 2 & MaxViT Block ×2 & 96 × 128 × 128 & 96 × 64 × 64 \\
& 3 & MaxViT Block ×2 & 96 × 64 × 64 & 192 × 32 × 32 \\
& 4 & MaxViT Block ×5 & 192 × 32 × 32 & 384 × 16 × 16 \\
& 5 & MaxViT Block ×2 & 384 × 16 × 16 & 768 × 8 × 8 \\
\midrule
\multirow{5}{*}{Skip Connection} 
& \multirow{2}{*}{2} & \multirow{2}{*}{CSI(CIU + SIU)} & 96 × 64 × 64 & \multirow{2}{*}{96 × 64 × 64} \\
& & & 96 × 64 × 64 \\
\cline{2-5}
& \multirow{2}{*}{3} & \multirow{2}{*}{CSI(CIU + SIU)} & 192 × 32 × 32 & \multirow{2}{*}{192 × 32 × 32} \\
& & & 192 × 32 × 32 \\
\cline{2-5}
& \multirow{2}{*}{4} & \multirow{2}{*}{CSI(CIU + SIU)} & 384 × 16 × 16 & \multirow{2}{*}{384 × 16 × 16} \\
& & & 384 × 16 × 16 \\
\midrule
\multirow{5}{*}{Decoder} & 1 & CoSE(Res(Conv3×3(stride=1) + BN + ReLU) + Res(SE)) & 96 × 128 × 128 & 48 × 128 × 128 \\
\cline{2-5}
& \multirow{2}{*}{2} 
  & CoSE(Res(Conv3×3(stride=1) + BN + ReLU) + Res(SE)) & \multirow{2}{*}{96 × 64 × 64}  & \multirow{2}{*}{96 × 128 × 128} \\
&   & + Bi(2×) + Conv3×3(stride=1) + BN + ReLU \\
\cline{2-5}
& \multirow{2}{*}{3} 
  & CoSE(Res(Conv3×3(stride=1) + BN + ReLU) + Res(SE)) & \multirow{2}{*}{192 × 32 × 32}  & \multirow{2}{*}{96 × 64 × 64} \\
&   & + Bi(2×) + Conv3×3(stride=1) + BN + ReLU \\
\cline{2-5}
& \multirow{2}{*}{4} 
  & CoSE(Res(Conv3×3(stride=1) + BN + ReLU) + Res(SE)) & \multirow{2}{*}{384 × 16 × 16}  & \multirow{2}{*}{192 × 32 × 32} \\
&   & + Bi(2×) + Conv3×3(stride=1) + BN + ReLU \\
\cline{2-5}
& \multirow{2}{*}{5} 
  & CoSE(Res(Conv3×3(stride=1) + BN + ReLU) + Res(SE)) & \multirow{2}{*}{768 × 8 × 8}  & \multirow{2}{*}{384 × 16 × 16} \\
&   & + Bi(2×) + Conv3×3(stride=1) + BN + ReLU \\
\midrule
\multirow{4}{*}{Output Module (MLF)} & & Conv1×1(stride=1) + CC  & 48 × 128 × 128 & \multirow{4}{*}{9 × 256 × 256} \\
& & + GroupConv3×3(stride=1) + BN + ReLU & 96 × 64 × 64 & \\
& & + Conv5×5(stride=1) & 192 × 32 × 32 & \\
& & & 384 × 16 × 16 & \\
\bottomrule
\end{tabular}
}
\label{architecture}
\end{table}

\subsection{CSI Module}

\begin{figure*}[!h]
    \centering
    \includegraphics[width=1\textwidth]{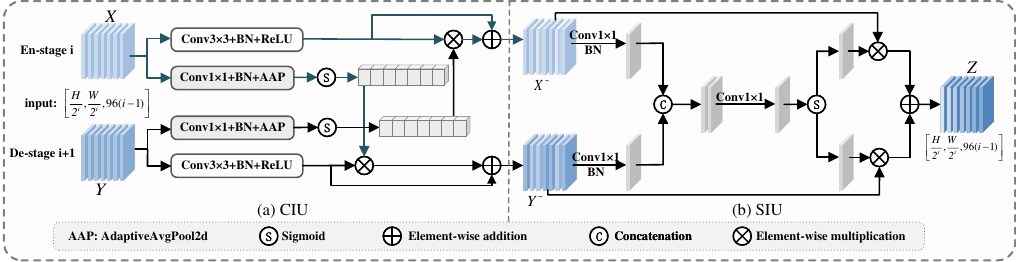}
    \caption{ The architecture of Channel Spatial Interaction Module.}
    \label{csi}
\end{figure*}

The CSI module is composed of two sequential sub-units: the Channel Interaction Unit (CIU) and the Spatial Interaction Unit (SIU), as shown in Figure \ref{csi}. The CSI is applied to the skip connections of the UNet architecture, where it takes as input the encoder features at stage $i$ ($i=2,3,4$) and the decoder features at stage $i+1$.

 The purpose is to improve the effectiveness of feature transfer across encoder and decoder levels by enhancing the semantic quality of skip connection features. Specifically:

\begin{itemize}
\item  The CIU is to align encoder and decoder features in the channel dimension by learning a set of adaptive channel weights, which is expected to emphasize semantically relevant channels. Since encoder and decoder features are often misaligned due to different levels of abstraction, the channel-wise recalibration also benefits to bridge the semantic gap.

\item  The SIU then complements the channel-level refinement by capturing spatial correlations between features, which enables the model to maintain spatial consistency and retain structural details to further support the generation of coherent segmentation outputs.
\end{itemize}

Finally, the enhanced feature map produced by the CSI module is fed into the decoder stage $i$. It is worth noting that the CSI module can be flexibly integrated into any UNet-like architecture, significantly improving the semantic expressiveness of skip connections and strengthening the decoder’s ability to utilize encoder features effectively.

\subsubsection{CIU}
As mentioned before, the inputs of the CIU are the feature maps of the encoder stage $i$ ($X$) and the decoder stage $i+1$ ($Y$). The core idea of the CIU is to adjust the importance of input features ($X$ or $Y$) based on another counterpart feature map ($Y$ or $X$), implemented by parallel paths for the inputs separately (Figure \ref{csi}(a)). 

To be specific, for each path, the 1x1 convolution and batch normalization operations are first performed to integrate the global information along the channel dimension, followed by the adaptive average pooling (AAP) operation to generate feature weights. 
The correlation weights $W_{x}$ and $W_{y}$ are obtained by sigmoid activation functions, which indicate the importance of different feature channels. The above process can be denoted by:
\begin{equation}
    {W_x}{\rm{  =  }}\sigma \left( {AAP\left( {BN\left( {Con{v_{1 \times 1}}\left( X \right)} \right)} \right)} \right)
\end{equation}
\vspace{-1.5\baselineskip}
\begin{equation}
    {W_y}{\rm{  =  }}\sigma \left( {AAP\left( {BN\left( {Con{v_{1 \times 1}}\left( Y \right)} \right)} \right)} \right)
\end{equation}
where \textit{AAP} denotes the adaptive average pooling operation, and $\sigma$ denotes the sigmoid activation.

Similarly, for each path, the input feature maps are further recalibrated by a block of (CNN, BN, and ReLU), which is subsequently fused by the learned correlation weights to reweight the importance of each channel to obtain the interacted feature maps.  
\begin{equation}
{X^1} = ReLU(BN(Con{v_{3 \times 3}}(X))
\end{equation}
\vspace{-1.5\baselineskip}
\begin{equation}
  {X^2} = {W_y} \times {X^1}
\end{equation}
\vspace{-1.5\baselineskip}
\begin{equation}
   {Y^1} = ReLU(BN(Con{v_{3 \times 3}}(Y))
\end{equation}
\vspace{-1.5\baselineskip}
\begin{equation}
    {Y^2} = {W_x} \times {Y^1}
\end{equation}

Finally, the residual mechanism is utilized to fuse the interacted feature maps while retaining the original feature inputs. 
The outputs of the CIU module are the feature maps ${X^ \sim }$ and ${Y^ \sim }$:
\begin{equation}
    {X^ \sim } = {X^1} + {X^2}
\end{equation}
\vspace{-1.5\baselineskip}
\begin{equation}
   {Y^ \sim } = {Y^1} + {Y^2}
\end{equation}

\subsubsection{SIU}
In general, the outputs of the CIU are fed into the SIU module to generate a fused feature map as the output of the CSI module. 
The SIU focuses on reweighting the importance of spatial pixels by an X-shaped path (as in Figure \ref{csi}(b)), where all feature channels share a single weight matrix.

To be specific, for the left part of the X-shaped path, the 1×1 convolution and BN operations are utilized to squeeze the channels of feature maps (${X^ \sim }$ and ${Y^ \sim }$) to 1, aiming to generate global contexts.  The concatenation operation is applied to generate an initial weight matrix by fusing both the encoder and decoder features along the channel dimension, as shown below:
\begin{equation}
  Q = Cat\left( {BN\left( {Con{v_{1 \times 1}}\left( {{X^ \sim }} \right)} \right), BN\left( {Con{v_{1 \times 1}}\left( {{Y^ \sim }} \right)} \right)} \right)
\end{equation}
where \textit{Cat} denotes the concatenation operation along the channel dimension.

In succession, the initial weight matrix $Q$ is further recalibrated by the 1×1 convolution operation to generate the sample-dependent weights, as in: 
\begin{equation}
  P  =  Con{v_{1 \times 1}}(Q)
\end{equation}

In the right part of the X-shaped path, the sigmoid activation function is performed to project the weight elements to $[0, 1]$, which is further performed on the SIU inputs to reweight their pixel-wise importance. Finally, the output feature map is formulated by addition operations, as in:
\begin{equation}
   Z  =  {X^ \sim } \times \sigma \left( P \right) + {Y^ \sim } \times \sigma \left( P \right)
\end{equation}

In summary, the sequential combination of CIU and SIU enables the CSI module to jointly optimize semantic relevance and spatial coherence, producing more informative and better-aligned feature representations.

\subsection{CoSE Module}

In the original UNet, each decoder stage consists of only two groups of (CNN, batch normalization (BN), and rectified linear unit (ReLU) activation).
The inference rules are illustrated below:
\begin{equation}
    {Z_1} = ReLU(BN(Con{v_{3 \times 3}}(Z)))
\end{equation}
\vspace{-1.5\baselineskip}
\begin{equation}
    {Z_2} = ReLU(BN(Con{v_{3 \times 3}}({Z_1})))
\end{equation}
where $Con{v_{K \times K}}$  denotes a convolution operation with a kernel size of \textit{K}.
\textit{BN} and \textit{ReLU} denote the batch normalization and rectified linear unit activation function, respectively.

 In medical images, target regions often share similar textures and shapes with complex background organs, further providing challenges to distinguish critical features. To address this, the channel attention SENet (\cite{Huetal2018}) is integrated into the decoder stages to model channel interdependencies. SENet adaptively recalibrates feature channels via learnable attention weights, enabling the network to extract informative channels related to the target regions while suppressing irrelevant background channels.

As shown in Figure \ref{overall}(a), the CoSE module is constructed by a convolution block and a SENet, in which corresponding residual connections are added before the convolution block and the SENet, respectively. 
The residual connection directly applies the addition operation to transmit the learned features into deeper layers, which can effectively solve the gradient vanishing and explosion by providing additional paths to enhance the information propagation. 
The mentioned process can be expressed as the equations:
\begin{equation}
    {Z_3} = {Z_2} + Z
\end{equation}
\vspace{-1.5\baselineskip}
\begin{equation}
    S = SENet({Z_3}) + {Z_3}
\end{equation}

 This design is intuitively motivated by enhancing channel-wise feature selection within the decoder to improve the representation of discriminative features, which is critical for highlighting target regions from complex backgrounds. The residual connections further ensure stable training and effective feature propagation.

\subsection{MLF Module}
In the proposed model, the pre-trained encoder extracts features from input images by progressively reducing spatial resolution while increasing channel depth.  The decoder then restores the spatial resolution to generate pixel-wise segmentation masks.
In the original UNet, up-sampling is typically performed using simple interpolation, which limits effective interactions between features of different scales and often results in loss of fine-grained details important for accurate segmentation.

\begin{figure*}[!htbp]
    \centering
    \includegraphics[width=1\textwidth]{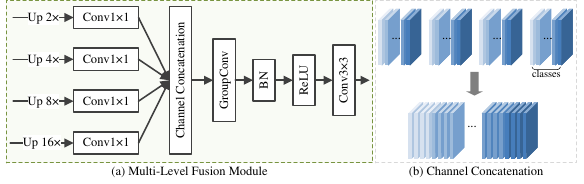}
    \caption{ (a) The architecture of MLF, (b) Process of Channel Concatenation.}
    \label{mlif}
\end{figure*}

 To address this issue,  a MLF module, as illustrated in Figure \ref{mlif}, is proposed to explicitly fuse features from different decoder stages at multiple scales. The MLF module enables richer feature interactions across scales through learnable fusion operations to preserve spatial details.
The goal of the MLF design is that integrating multi-scale information allows the network to combine coarse semantic context with fine spatial details, resulting in more precise and robust segmentation outputs.

The output feature maps from decoder stages 1-4 serve as the MLF inputs, which are up-sampled by 2, 4, 8, and 16 to generate image size-liked masks.
Then, the channel number of the up-sampled feature maps is mapped to the class number relevant to the task by a point-wise convolution. The inference rules are listed as:
\begin{equation}
  {Y_i}{\rm{  =  }}Con{v_{1 \times 1}}(U{p_{j \times }}({X_i})),  i = 1,2,3,4. j = 2,4,8,16.
\end{equation}
where ${X_i}$ denotes the output of the decoder stage $i$, and $U{p_{j \times }}$ denotes up-sampling ${X_i}$ by a factor of $j$. ${Y_i} \in R{^{N \times C \times H \times W}}$, where $N$ is the batch size, $C$ is the class number, $H$ and $W$ are the height and width of the original image, respectively.
In this process, each feature channel is regarded as the feature representation for the corresponding class, which guides the model to capture discriminative features for certain classes of the segmentation targets. 

To further optimize the feature representations, the corresponding channels of ${Y_1}$, ${Y_2}$, ${Y_3}$ and ${Y_4}$ are concatenated to combine the features of each class, as in Figure \ref{mlif}(b). 
\begin{equation}
  {Y_c} = ChannelCat({Y_1},{Y_2},{Y_3},{Y_4})
\end{equation}
where \textit{ChannelCat} denotes the channel concatenation operation.

Finally, a two-step convolution block is employed to integrate intra- and inter-class features to generate the segmentation mask.  
\begin{itemize}
    \item Intra-class: the group convolution performs convolution operations on feature channels for each class separately by setting the group number to the class number, which generates the feature maps with channel 1 for each class.
    \item Inter-class: the standard convolution is then used to integrate the learned intra-class features to obtain the final segmentation mask by fusing the intra-class features and inter-class features.
\end{itemize}
\begin{equation}
  G = ReLU(BN(GCon{v_{3 \times 3}}({Y_c}) ))
\end{equation}
\vspace{-1.5\baselineskip}
\begin{equation}
  output {\rm{ =  }}Con{v_{5 \times 5}}(G)
\end{equation}
where ${Y_c} \in R{^{N \times 4C \times H \times W}}$, $G \in R{^{N \times C \times H \times W}}$, $GCon{v_{3 \times 3}}$ denotes the group convolution with a kernel size of 3.

\subsection{Loss Function}
Based on the segmentation mask, the loss function of the FIF-UNet is obtained by a weighted Dice loss and the Cross-Entropy (CE) loss, which are designed for segmentation and classification tasks, as in: 
\begin{equation}
 L = {\lambda _1}{L_{DICE}} + {\lambda _2}{L_{CE}}
\label{loss}
\end{equation}
where ${\lambda _1}$ and ${\lambda _2}$ are the weights for the Dice loss (${L_{DICE}}$) and CE loss (${L_{CE}}$), respectively.

To enhance the model convergence, the multi-stage feature mixing loss aggregation (MUTATION) method proposed by MERIT (\cite{RahmanMarculescu2024}) is introduced in this work. 
To be specific, for the feature maps ($Y_1$, $Y_2$, $Y_3$, $Y_4$) from the MLF module, a total of 15 ($2^4-1$) nonempty subsets are first obtained, based on which 15 predicted masks are generated by element-wise addition on the feature maps in each set. 
Consequently, for each mask in the 15 predictions, the $L$ in Equation \ref{loss} is calculated over the ground truth to measure multi-stage prediction errors. 
In addition, the loss between the MLF output and the ground truth is also considered to formulate the final loss, as in:
\begin{equation}
Sets  =  nonsubset([{Y_1},{Y_2},{Y_3},{Y_4}])
\end{equation}
\vspace{-1.5\baselineskip}
\begin{equation}
R[i] = \sum\nolimits_{j = 0}^n {Sets[i][j]} ,i = 0,1,\cdots,14 
\end{equation}
\vspace{-1.5\baselineskip}
\begin{equation}
loss = L(output,label) + \sum\nolimits_{i = 0}^{14} {L(R[i],label)} 
\end{equation}
where $nonsubset$ denotes a function that takes nonempty subsets of a list. $n$ denotes the element number in each subset.

\section{Experiments and results}
\subsection{Datasets and evaluation metrics}
\textbf{Synapse multi-organ segmentation dataset}: 30 abdominal CT scans are included in the Synapse (\cite{Landmanetal2015}), with a total of 3779 axial contrast-enhanced abdominal CT images, which is provided by the MICCAI 2015 Multi-Atlas Abdomen Labeling Challenge.
Each CT scan consists of 85-198 slices of 512 × 512 pixels, and the voxel spatial resolution is ([0:54-0:54] × [0:98-0:98] × [2:5-5:0])$m{m^3}$. 
 Following the common practices in previous works, the dataset is randomly divided into 18 scans (2212 axial slices) for training, and 12 for validation. Notably, the 12 evaluation scans are commonly referred to as the test set in existing studies and are used for reporting final performance. In this work, the same setting is adopted for fair comparison among baselines.
 During the model training, the Dice score (DICE) serves as the optimization metric, while during the evaluation, both Dice score (primary metric) and 95\% Hausdorff Distance (HD95) are computed to present a comprehensive performance.
A total of 8 anatomical structures are segmented on Synapse, including the aorta, gallbladder, left kidney, right kidney, liver, pancreas, spleen, and stomach. 

\begin{table*}[]
 \caption{ Comparative analysis of model performance on Synapse dataset for organ segmentation. Organ abbreviations: GB (gallbladder), KL (left kidney), KR (right kidney), PC (pancreas), SP (spleen), SM (stomach). Only Dice scores are reported for individual organs. High Dice scores and low HD95 scores mean better performance. The best result is highlighted in bold, and the second-best is highlighted with an underline. }
\centering
\resizebox{1\textwidth}{!}{
\begin{tabular}{c|c|cc|cccccccc|cc}
\toprule
\multirow{2}{*}{} & \multirow{2}{*}{Methods} & \multicolumn{2}{c|}{Average} & \multirow{2}{*}{Aorta} & \multirow{2}{*}{GB} & \multirow{2}{*}{KL} & \multirow{2}{*}{KR} & \multirow{2}{*}{Liver} & \multirow{2}{*}{PC} & \multirow{2}{*}{SP} & \multirow{2}{*}{SM} & Params & FLOPs\\
                  &   & DICE↑        & HD95↓        &                        &                     &                     &                     &                        &                     &                     &                    &(M) &(G) \\
\midrule
\multirow{2}{*}{CNN}& UNet           & 70.11                                   & 44.69                                   &                    84.00 & 56.70 & 72.41 & 62.64 & 86.98 & 48.73 & 81.48 & 67.96  & 31.04 & 54.8 \\
& AttnUNet       & 71.70                                   & 34.47                                            & 82.61 & 61.94 & 76.07 & 70.42 & 87.54 & 46.70 & 80.67 & 67.66 & 34.88 & 66.64 \\
& nnU-Net       & 82.63                                   & -                                            & \textbf{91.28} & 64.06 & 83.57 & 81.31 & 94.53 & \underline{73.67} & 88.60 & 84.05 & 31.06 & 78.38 \\
\midrule
\multirow{3}{*}{ViT}& SwinUNet       & 79.13                                   & 21.55                                   & 85.47 & 66.53 & 83.28 & 79.61 & 94.29 & 56.58 & 90.66 & 76.60 & 27.17 & 5.92 \\
& TransDeepLab   & 80.16                                   & 21.25                                   & 86.04 & 69.16 & 84.08 & 79.88 & 93.53 & 61.19 & 89.00 & 78.40 & 21.14 & 15.75 \\
& MISSFormer     & 81.96                                   & 18.20                                   & 86.99 & 68.65 & 85.21 & 82.00 & 94.41 & 65.67 & 91.92 & 80.81 & 42.46 & 7.28 \\
\midrule
\multirow{11}{*}{Hybrid}& TransUNet      & 77.48                                   & 31.69                                   & 87.23 & 63.13 & 81.87 & 77.02 & 94.08 & 55.86 & 85.08 & 75.62 & 105.28 & 25.41 \\
& SSFormerPVT    & 78.01                                   & 25.68                                   & 82.78 & 63.74 & 80.72 & 78.11 & 93.53 & 61.53 & 87.07 & 76.61 & - & - \\
& PolypPVT       & 78.08                                   & 25.61                                   & 82.34 & 66.14 & 81.21 & 73.78 & 94.37 & 59.34 & 88.05 & 79.40 & 25.11 & 10.10 \\
& MT-UNet        & 78.59                                   & 26.59                                   & 87.92 & 64.99 & 81.47 & 77.29 & 93.06 & 59.46 & 87.75 & 76.81 & 79.08 & 44.87 \\
& HiFormer       & 80.29                                   & 18.85                                   & 85.63 & 73.29 &             82.39 & 64.84 & 94.22 & 60.84 & 91.03 & 78.07 &25.51 & 13.33 \\
& PVT-CASCADE    & 81.06                                   & 20.23                                   & 83.01 & 70.59 & 82.23 & 80.37 & 94.08 & 64.43 & 90.10 & 83.69 & 35.28 & 7.62 \\
& CASTformer     & 82.55                                   & 22.73                                   & 89.05 & 67.48 & 86.05 & 82.17 & \underline{95.61} & 67.49 & 91.00 & 81.55 & - & - \\
& TransCASCADE   & 82.68                                   & 17.34                                   & 86.63 & 68.48 & 87.66 & 84.56 & 94.43 & 65.33 & 90.79 & 83.52 & 123.49  & 41.03 \\
& Cascaded MERIT & \underline{84.90}                                   & \textbf{13.22}                                   & 87.71 & 74.40 & \underline{87.79} & \underline{84.85} & 95.26 & 71.81 & \underline{92.01} & 85.38 & 147.86 & 33.44 \\
 &Tiny FIF-UNet(ours)  & 84.40                                   & 19.89                                   & 89.19 & \underline{74.56} & 85.22 & 81.39 & 95.19 & 70.69 & \textbf{92.11} & \underline{86.42} & 38.31 & 11.21 \\
&Small FIF-UNet(ours) & \textbf{86.05}                                   & \underline{15.82}                                   & \underline{89.49} & \textbf{76.15} & \textbf{88.23} & \textbf{86.26} & \textbf{95.87} & \textbf{74.14} & 91.31 & \textbf{86.97} & 86.91 & 23.26 \\ 
\bottomrule
\end{tabular}
}
\label{synapse}
\end{table*}

\textbf{Automated cardiac diagnosis challenge}: The ACDC dataset (\cite{Bernardetal2018}) consists of 100 cardiac MRI scans collected from different patients, provided by the MICCAI ACDC challenge 2017. 
Each scan contains three organs: the right ventricle, left ventricle, and myocardium. 
 Following the official data split in the ACDC challenge and consistent with previous works, the dataset is divided into 70 cases (1930 axial slices) for training, 10 for validation, and 20 for testing. The results reported in this paper are based on the test set to ensure fair comparison and reproducibility.
 The Dice score is used consistently as the evaluation metric during all stages of training, validation, and testing on this dataset.

\subsection{Implementation details}
The pre-trained MaxViT from (\cite{Tuetal2022}) serves as the encoder of the proposed model, with the input resolution of 256×256 and attention window size of 7×7. 
To consider the final performance, both the small and tiny MaxViT architectures are applied to conduct the experiments, i.e., Small FIF-UNet and Tiny FIF-UNet.
 To enhance the diversity of the training samples, we implement three data augmentation strategies: (1) discrete random rotation (90°, 180°, or 270°), (2) random flipping (horizontal or vertical), and (3)  continuous random rotation (with angles varying within ±20°). 
The model is trained using AdamW (\cite{LoshchilovHutter2017}) optimizer with the learning rate of 1e-4 for 400 epochs, applying the weight decay of 1e-4. 
The batch size of 16 is used for Synapse and ACDC. 
Following (\cite{Tuetal2022}), the loss weights ${\lambda _1}$ and ${\lambda _2}$ are set to 0.7 and 0.3, respectively. 
The proposed model is implemented using Pytorch 2.2.2 and all experiments are conducted on a single NVIDIA TITAN RTX GPU with 24GB of memory. 

\subsection{Quantitative Results}
This section provides quantitative comparisons of the proposed model against state-of-the-art methods in terms of both segmentation accuracy and computational efficiency. Accuracy is evaluated on the primary Synapse dataset and the ACDC dataset. Computational efficiency is measured and compared using model parameter count (Params) and floating-point operations (FLOPs). The results validate that our model achieves superior accuracy, robust generalization, and a better trade-off between performance and resource consumption. The detailed quantitative results are presented below.

\subsubsection{Experimental results on Synapse dataset}
The experimental results on the Synapse multi-organ dataset are reported in Table \ref{synapse}, including the proposed model and other selective baselines.
A total of 14 comparative baselines are selected to evaluate the model performance, as in three categories, CNN-based models (i.e., UNet (\cite{Ronnebergeretal2015}), AttnUNet (\cite{Oktayetal2018}), nnU-Net(\cite{isensee2021nnu})), ViT-based models (i.e., SwinUNet (\cite{Caoetal2022}), TransDeepLab (\cite{Azadetal2022}), MISSFormer (\cite{Huangetal2021})) and hybrid CNN-Transformer models (i.e., TransUNet (\cite{Chenetal2021}), SSFormerPVT (\cite{Wangetal2022}), PolypPVT (\cite{Dongetal2021}), MT-UNet (\cite{Wangetal2022}), HiFormer (\cite{Heidarietal2023}), PVT-CASCADE (\cite{RahmanMarculescu2023}), CASTformer (\cite{Youetal2022}), TransCASCADE (\cite{RahmanMarculescu2023}), Cascaded MERIT (\cite{RahmanMarculescu2024})). 

As shown in Table \ref{synapse}, the average DICE and HD95 are reported to compare the model performance, as well as the Dice score for the certain 8 classes. 
In general, the Small FIF-UNet achieves the highest average DICE of 86.05\% (primary metric), which significantly outperforms all the selective baselines (1.15\% absolute improvement over the best baseline).
Specifically, the Small FIF-UNet has the ability to harvest the best performance for 7/8 classes, confirming the performance superiority over baselines.
Meanwhile, the proposed model also obtains the second-best HD95 measurement. Compared to classical ViT-based SwinUNet and hybrid TransUNet, the Small FIF-UNet improves the performance in terms of average DICE by 6.92\% and 8.57\%, respectively. 
 
  More importantly, the proposed FIF-UNet demonstrates significant performance improvements, particularly in small and hard-to-segment organs. Compared to the Cascaded MERIT(use the same pre-trained encoder), on two small and relatively underrepresented organs, the Small FIF-UNet achieves improvements of 0.44\% on the left kidney (KL) and 1.41\% on the right kidney (KR), which can be attributed that the MLF module can enhance feature detail through multi-scale fusion to improve the model’s ability to detect small structures.

 For challenging organs with low contrast and ambiguous boundaries, such as the gallbladder (GB) and pancreas (PC), the proposed model achieves even larger performance gains of 1.75\% and 2.33\%, respectively. These improvements benefit from the combined effects of the CSI module to strengthen semantic interactions between encoder and decoder layers by preserving more informative details, and the CoSE module to enhance the model’s capacity to represent discriminative features. By combining the two blocks, the model provides better anatomical consistency across organs of varying size and difficulty.

 In addition, the proposed Tiny FIF-UNet and Small FIF-UNet achieve an excellent balance between segmentation accuracy and computational efficiency. From the efficiency perspective, the Tiny FIF-UNet achieves an average Dice score of 84.40\% with only 38.31M parameters and 11.21 GFLOPs, outperforming models with comparable or even higher complexity (i.e., AttnUNet (71.70\%, 34.88M, 66.64G) and HiFormer (80.29\%, 25.51M, 13.33G)), demonstrating its superior accuracy-efficiency performance. From the accuracy perspective, the Small FIF-UNet achieves the highest average Dice score of 86.05\%, outperforming all baselines including TransCASCADE (82.68\%, 123.49M, 41.03G) and Cascaded MERIT (84.90\%, 147.86M, 33.44G), while using significantly fewer parameters and FLOPs. The results show that the proposed model can provide better performance at a lower computational cost.

In summary, the experimental results demonstrate that the proposed model achieves both performance and efficiency superiority over selected baselines on the Synapse dataset, which can also harvest desired enhancement on small and hard-to-segment organs.

\subsubsection{Experimental results on ACDC dataset}
\begin{table}[]
\caption{ Comparative analysis of model performance on ACDC dataset for organ segmentation. Organ abbreviations: RV (right ventricle), Myo (myocardium), LV (left ventricle). Only Dice scores are reported for individual organs. The best result is highlighted in bold, and the second-best is highlighted with an underline.}
\centering
\begin{tabular}{c|c|c|ccc|cc}
\toprule
         & Methods  & Avg DICE↑ & RV    & Myo   & LV    & Params(M)  & FLOPs(G) \\
\midrule
\multirow{2}{*}{CNN}& R50 AttnUNet   & 86.75    & 87.58 & 79.20 & 93.47 & - & - \\
                    & R50 UNet       & 87.55    & 87.10 & 80.63 & 94.92 & - & -\\
                     & nnU-Net       & 91.59    & 90.53 & 89.91 & 94.33 & 31.06 & 78.38 \\
\midrule
\multirow{2}{*}{ViT}& SwinUNet       & 90.00    & 88.55 & 85.62 & 95.83 & 27.17 & 5.92 \\
                       & MISSFormer     & 90.86    & 89.55 & 88.04 & 94.99 & 42.46  & 7.28 \\
\midrule
\multirow{8}{*}{Hybrid}   & TransUNet      & 89.71    & 88.86 & 84.53 & 95.73 & 105.28 & 25.41 \\
& MT-UNet      & 90.43    & 86.64 & 89.04 & 95.62 & 79.08  & 44.87 \\
& PVT-CASCADE    & 91.46    & 88.90 & 89.97 & 95.50 & 35.28 & 7.62\\
& TransCASCADE   & 91.63    & 89.14 & \textbf{90.25} & 95.50 & 123.47 & 41.03\\
& Cascaded MERIT & 91.85    & 90.23 & 89.53 & 95.80 & 147.86 & 33.44 \\
& Parallel MERIT & 92.32    & 90.87 & 90.00 & \underline{96.08} & 147.86 & 33.44 \\
& Tiny FIF-UNet(ours)  & \underline{92.37}        & \underline{91.04}     & 90.10     & 95.96     & 38.31 & 11.21 \\
& Small FIF-UNet(ours) & $\mathbf{92.58 \pm 0.12}$        & \textbf{91.30}     & \underline{90.24}     & \textbf{96.19}  
& 86.91 & 23.26 \\
\bottomrule
\end{tabular}
\label{acdc}
\end{table}

 The experimental results of comparative methods on the ACDC dataset are reported in Table \ref{acdc}, in terms of the average Dice score.
Similarly, three categories of models are selected as the comparative baselines, including  CNN-based models (i.e., R50 UNet (\cite{Chenetal2021}), R50 AttnUNet (\cite{Chenetal2021}), nnU-Net(\cite{isensee2021nnu})), ViT-based models (i.e., SwinUNet (\cite{Caoetal2022}), MISSFormer (\cite{Huangetal2021})) and hybrid CNN-Transformer models (i.e., TransUNet (\cite{Chenetal2021}), MT-UNet (\cite{Wangetal2022}), PVT-CASCADE (\cite{RahmanMarculescu2023}), TransCASCADE (\cite{RahmanMarculescu2023}), Cascaded MERIT (\cite{RahmanMarculescu2024}), Parallel MERIT (\cite{RahmanMarculescu2024})).

 As shown in Table \ref{acdc}, the proposed Small FIF-UNet achieves an average Dice score of 92.58±0.12\% on the ACDC dataset across eight independent runs, with a confidence range of 0.12. Compared to the best baseline Parallel MERIT (92.32\%) with the same pre-trained encoder, the proposed model also achieves 0.26\% improvement, and the tight confidence range clearly indicates the greater stability and reliability of our model.

 Specifically, the Small FIF-UNet achieves the highest Dice scores on RV (91.30\%) and LV (96.19\%), and comparable performance for Myo (90.24\%), only 0.01\% below the best method. These improvements are especially valuable given the challenges of segmenting the right ventricle and myocardium, which often exhibit blurry boundaries and low contrast with the background.

 It is believed that the consistent performance improvement can be attributed to the proposed architectural components: the CSI module enhances multi-scale semantic fusion via skip connections; CoSE improves the focus on discriminative regions;   and MLF reinforces fine-grained structural details. By integrating them into the proposed model, both feature representation and boundary localization can be enhanced and results across Synapse and ACDC datasets also demonstrate the robust generalization capability  across different data modalities.

 Similarly, the proposed model achieves the balance between segmentation accuracy and computational efficiency to the Synapse dataset, as that in ACDC. Notably, the proposed Tiny FIF-UNet achieves the second-best average Dice score of 92.37\%, closely matching Parallel MERIT (92.32\%) while using only 38.31M trainable parameters v.s. 109.55M in Parallel MERIT. Considering that the two models adopt the same pre-trained encoder, this result further validates the strong capability of the proposed modules to enhance feature representation and segmentation performance with significantly lower
computational cost.

\subsection{Ablation studies}
To comprehensively evaluate the effectiveness of each proposed component, we conduct a series of ablation studies on two benchmark datasets.   Specifically, we first analyze the contribution of each module (CSI, CoSE, and MLF) on the Synapse dataset, followed by a similar investigation on the ACDC dataset to verify the generalization ability across different tasks.   In addition, we compare different loss functions to assess their impact on segmentation performance.   Finally, we evaluate the superiority of our CSI module by comparing it with alternative skip connection enhancement strategies.  
\subsubsection{Ablation experiment on Synapse dataset}
\begin{table*}[]
\caption{Ablation studies based on the Synapse dataset}
\centering
\resizebox{1\textwidth}{!}{
\begin{tabular}{llc|lc|llllllll}
\toprule
\multirow{2}{*}{CoSE} & \multirow{2}{*}{CSI} & \multirow{2}{*}{MLF} & \multicolumn{2}{c|}{Average} & \multirow{2}{*}{Aorta} & \multirow{2}{*}{GB} & \multirow{2}{*}{KL} & \multirow{2}{*}{KR} & \multirow{2}{*}{Liver} & \multirow{2}{*}{PC} & \multirow{2}{*}{SP} & \multirow{2}{*}{SM} \\
            &      &   & DICE↑        & HD95↓        &                        &                     &                     &                     &                        &                     &                     &                     \\
\midrule
 $\times$   & $\times$  & $\times$     & 84.57  & 18.58 & 88.61 & 75.78 & 84.87 & 82.74 & 95.31 & 72.60 & 91.81 & 84.81  \\
\midrule
 \checkmark    & $\times$    & $\times$     & 85.47   & 19.50 & 89.28 & 74.83 & 86.86 & 85.12 & 95.05 & 75.04 & 91.46 & 86.11\\
 $\times$   & \checkmark    & $\times$    & 85.46   & 15.37 & 90.06 & 73.51 & 88.31& 85.36 & 95.30 & 73.45 & 91.12& 86.55\\
 $\times$   & $\times$    & \checkmark     & 85.45   & 12.95 & 89.07 & 76.80 & 88.80 & 85.73 & 95.14 & 72.62 & 90.73 & 84.68\\
\midrule
 \checkmark    & \checkmark    & $\times$     & 85.92   & 15.33 & 89.27 & 76.74 & 87.43 & 86.05 & 95.57 & 75.02 & 92.06 & 85.25\\
 \checkmark    & $\times$    & \checkmark     & 85.57   & 14.08 & 88.87 & 74.47 & 88.71 & 85.36 & 95.40 & 73.57 & 93.02 & 85.13\\
$\times$    & \checkmark    & \checkmark     & 85.90   & 11.34 & 88.50 & 74.41 & 88.62 & 86.39 & 95.76 & 72.66 & 93.86 & 87.04 \\
\midrule
 \checkmark    & \checkmark    & \checkmark     & 86.05   & 15.82                                   & 89.49 & 76.15 & 88.23 & 86.26 & 95.87 & 74.14 & 91.31 & 86.97 \\
\bottomrule
\end{tabular}
}
\label{ablation}
\end{table*}

 To validate the effectiveness and complementarity of each proposed module, we conducted ablation studies on the Synapse dataset, as summarized in Table \ref{ablation}. The experimental design for ablation studies concerns the employment of the proposed technical modules separately or their combinations with the baseline.

 It can be seen that the CoSE module enhances the model’s ability to capture discriminative features, which is particularly beneficial for segmenting organs with low contrast to the background or ambiguous anatomical boundaries. Specifically, the CoSE improves the Dice scores for the SM and PC by 1.30\% and 2.44\%, respectively. The CSI module facilitates semantic fusion between encoder and decoder features to effectively preserve more contextual and detailed information, achieving improvements for both small and ambiguous organs, with gains of 3.44\% and 2.62\% on KL and KR, and 1.74\% on the SM. The MLF module strengthens fine-grained feature representations through multi-level fusion to support the segmentation of small-scale structures, harvesting the most significant performance improvements for KL and KR by 3.93\% and 2.99\%, respectively.

 In addition, by combining two of them into the proposed model, distinct advantages are observed depending on organ characteristics. To be specific, the combination of CSI and CoSE yields notable improvements for challenging organs such as the GB and PC, with Dice score increases of 0.96\% and 2.42\%, respectively. These results suggest that CSI's semantic enhancement and CoSE's discriminative focus are complementary. On the other hand, CSI+MLF and CoSE+MLF combinations show greater improvements for small organs, as they can preserve and refine detailed information at different stages of the network.

 Finally, the full FIF-UNet model, incorporating all three modules, achieves the best overall performance with an average Dice score of 86.05\%, representing a 1.48\% improvement over the baseline. It is believed that the comprehensive enhancement across small, ambiguous, and difficult organs can validate the proposed modules and the overall effectiveness of the proposed architecture.

\subsubsection{Ablation experiment on ACDC dataset}
\begin{table*}[!h]
\caption{Ablation studies based on the ACDC dataset}
\centering
\begin{tabular}{ccccccc}
\toprule
 CoSE & CSI & MLF  & Avg DICE↑ & RV    & Myo   & LV \\
\midrule
 $\times$   & $\times$  & $\times$     & 91.94  & 89.96 & 90.01 & 95.84  \\
\midrule
 \checkmark    & $\times$    & $\times$     & 92.27   & 90.83 & 90.02 & 95.95 \\
 $\times$   & \checkmark    & $\times$    & 92.30   & 90.79 & 90.25 & 95.86 \\
 $\times$   & $\times$    & \checkmark     & 92.26   & 90.70 & 89.93 & 96.15 \\
\midrule
 \checkmark    & \checkmark    & $\times$     & 92.47   & 90.54 & 90.47 & 96.39 \\
 \checkmark    & $\times$    & \checkmark     & 92.46   & 91.13 & 90.22 & 96.02 \\
$\times$    & \checkmark    & \checkmark     & 92.51   & 91.07 & 90.32 & 96.13  \\
\midrule
 \checkmark    & \checkmark    & \checkmark    &  92.58 & 91.30 & 90.24 & 96.19  \\
\bottomrule
\end{tabular}

\label{ablacdc}
\end{table*}
Table \ref{ablacdc} presents ablation results on the ACDC dataset, evaluating the effectiveness and complementarity of the proposed CoSE, CSI, and MLF modules in cardiac structure segmentation. As can be seen from the results, each proposed  module, whether independent employment
or in combination, can also contribute desired performance improvements in this work.

 For the individual modules, the CoSE module results in the most notable improvement in the segmentation of the RV (the Dice score increases by 0.87\%), which confirms that CoSE enhances the model's ability to capture discriminative features, particularly in regions with low contrast and ambiguous boundaries. The CSI module, designed to facilitate semantic interactions between encoder and decoder features, consistently improves performance across all three anatomical structures, increasing the average Dice score by 0.36\%, which highlights CSI’s effectiveness in preserving and transmitting semantic information. The MLF module, which performs multi-scale feature fusion at the output stage, significantly boosts segmentation performance for the LV with a relatively regular shape and well-defined boundaries, which validates MLF’s strength in enhancing structural detail.

 Similarly, further performance gains are observed for evaluating pairwise combinations of the modules. The integration of CSI and CoSE produces significant improvements for challenging structures such as the RV and Myo, with Dice scores reaching 90.54\% and 90.47\%, respectively, and an average Dice gain of 0.53\%. The results suggest that combining discriminative feature enhancement with multi-level semantic interactions is particularly effective for segmenting complex or ambiguous anatomical regions. Similarly, the combination of CoSE and MLF improves RV and Myo segmentation with an average Dice gain of 0.52\%, indicating that enhancing boundary details further strengthens discriminative feature representations. Although the CSI and MLF combination lacks the explicit discriminative modeling provided by CoSE, it still achieves the highest average Dice improvement (0.57\%) among the pairwise configurations, due to effective integration of semantic context and structural detail at different stages of the network.

 The full model FIF-UNet with all three modules achieves the best performance (average Dice: 92.58\%, ↑ 0.64\%), confirming their complementary roles and strong synergy in improving segmentation accuracy and robustness across all cardiac structures.

\subsubsection{Comparison of different loss functions}
\begin{table*}[]
\caption{ Comparative analysis of model performance based on different loss functions on Synapse dataset for organ segmentation. Organ abbreviations: GB (gallbladder), KL (left kidney), KR (right kidney), PC (pancreas), SP (spleen), SM (stomach). Only Dice scores are reported for individual organs. High Dice scores and low HD95 scores mean better performance. The best result is highlighted in bold, and the second-best is highlighted with an underline.}
\centering
\resizebox{1\textwidth}{!}{
\begin{tabular}{cc|lc|llllllll}
\toprule
 \multirow{2}{*}{loss1} &\multirow{2}{*}{loss2} & \multicolumn{2}{c|}{Average} & \multirow{2}{*}{Aorta} & \multirow{2}{*}{GB} & \multirow{2}{*}{KL} & \multirow{2}{*}{KR} & \multirow{2}{*}{Liver} & \multirow{2}{*}{PC} & \multirow{2}{*}{SP} & \multirow{2}{*}{SM} \\
  & & DICE↑        & HD95↓        &                        &                     &                     &                     &                        &                     &                     &                     \\
\midrule

Generalized DICE &CE Loss          & 84.36       & \textbf{15.24}            & \textbf{89.76} & 69.21 & \textbf{88.77} & 84.69 & 95.37 & 72.26 & 90.31 & 84.54 \\
Tversky Loss &CE Loss       & 85.35  & 24.88    & 88.56                         & \textbf{76.78} & 86.48 & 84.09 & \underline{95.50} & \textbf{74.82} & 90.83 & \underline{85.77}  \\
 Dice Loss &Focal Loss       & \underline{85.42}                                   & 17.28                                   & 88.60 & \underline{76.36} & 87.62 & \underline{84.86} & 95.29 & 74.11 & \underline{91.08} & 85.42 \\

 \midrule
 Dice Loss &CE Loss   & \textbf{86.05}                                   & \underline{15.82}                                   & \underline{89.49} & 76.15 & \underline{88.23} & \textbf{86.26} & \textbf{95.87} & \underline{74.14} & \textbf{91.31} & \textbf{86.97} \\

\bottomrule
\end{tabular}
}
\label{loss}
\end{table*}
\begin{figure*}[!htbp]
    \centering
    \includegraphics[width=1\textwidth]{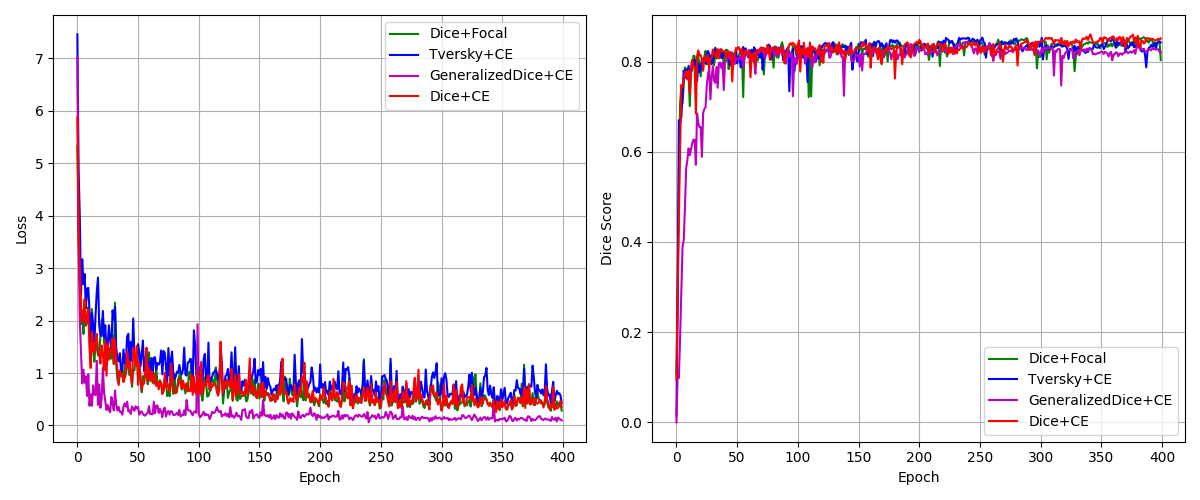}
    \caption{ Comparison of training loss and Dice score curves under different loss function combinations. Dice + CE Loss achieves the most stable convergence and the highest segmentation performance.}
    \label{lossv}
\end{figure*}
 To evaluate the impact of different loss functions on segmentation performance, we perform a comparative study on the Synapse dataset, as shown in Table \ref{loss}. Among all configurations, the combination of Dice Loss and CE Loss achieves the best overall performance, yielding the highest average Dice score (86.05\%) and the second-lowest HD95 distance (15.82). In addition, this combination produces either the best or second-best Dice scores for most individual organs, indicating both high segmentation accuracy and strong anatomical consistency, which highlights the effectiveness and robustness of this loss formulation for medical image segmentation tasks.

 The superior performance of the Dice + CE combination benefits from the complementary properties of the two loss functions. Dice Loss effectively addresses class imbalance by directly optimizing region overlap, which is especially beneficial for small or underrepresented organs. However, it often produces unstable gradients during early training. In contrast, Cross-Entropy Loss provides stable pixel-wise supervision, enables faster convergence, and improves boundary delineation in the initial stages. By combining them to perform the model training, CE Loss guides the model in capturing coarse structural features, while Dice Loss enhances global region alignment, resulting in improved segmentation accuracy and anatomical consistency.

 Other loss function combinations offer certain advantages in specific scenarios but fail to achieve comparable overall performance. Generalized Dice Loss, although more suitable for extreme class imbalance, introduces additional complexity and typically yields lower Dice scores across most organs. Tversky Loss, designed to balance false positives and false negatives, depends heavily on task-specific tuning and fails to generalize across diverse classes. Focal Loss focuses on hard-to-classify pixels but lacks the region-level structural sensitivity provided by CE Loss and further leads to reduced robustness. These comparisons further confirm the effectiveness and generalization capability of the Dice + CE combination in multi-organ segmentation tasks.

 To further investigate the impact of different loss function combinations on training loss and segmentation performance, the training convergence is shown in Figure \ref{lossv}. Among all configurations, the combination of Dice Loss and CE Loss exhibits the most stable and efficient training. The average loss decreases rapidly with minimal fluctuations and stabilizes early, while the Dice score reaches 0.8 within the first 50 epochs and remains consistently high results throughout the training process. In contrast, the Dice + Focal Loss demonstrates a similar initial convergence pattern but suffers from greater fluctuations in both the loss and Dice score, i.e., unstable performance.

 The Tversky Loss + CE Loss combination results in poor convergence efficiency with unstable loss values. Although the Dice score increases relatively fast in the early stages, it can only obtain inferior final results with limited overall learning capacity. The Generalized Dice + CE Loss shows the fastest decline in loss, prematurely approaching zero and further to overfitting. Meanwhile, the resulting Dice score improves slowly and remains the lowest among all configurations. In summary, these observations confirm that the Dice + CE Loss combination achieves the best balance between fast convergence, stable optimization, and final segmentation accuracy.

\subsubsection{Comparison of different skip connection enhancement methods}

\begin{table*}[]
\caption{ Comparative analysis of model performance based on different skip connection enhancement methods on Synapse dataset for organ segmentation. Organ abbreviations: GB (gallbladder), KL (left kidney), KR (right kidney), PC (pancreas), SP (spleen), SM (stomach). Only Dice scores are reported for individual organs. High Dice scores and low HD95 scores mean better performance. The best result is highlighted in bold, and the second-best is highlighted with an underline.}
\centering
\resizebox{1\textwidth}{!}{
\begin{tabular}{c|lc|llllllll}
\toprule
 \multirow{2}{*}{Methods} & \multicolumn{2}{c|}{Average} & \multirow{2}{*}{Aorta} & \multirow{2}{*}{GB} & \multirow{2}{*}{KL} & \multirow{2}{*}{KR} & \multirow{2}{*}{Liver} & \multirow{2}{*}{PC} & \multirow{2}{*}{SP} & \multirow{2}{*}{SM} \\
  & DICE↑        & HD95↓        &                        &                     &                     &                     &                        &                     &                     &                     \\
\midrule

 AG           & 84.06       & 17.78            & 89.05 & 71.21 & 86.56 & 83.29 & 94.96 & 73.92 & 90.74 & 82.77 \\
 FFM       & 85.15  & 18.77    & 88.88                         & 73.74 & 85.11 & 84.16 & 95.40 & 73.71 & \underline{92.96} & \textbf{87.24}  \\
 GAB       & 85.16                                   & 15.83                                   & 88.42 & 74.62 & \underline{89.56} & 84.91 & 94.99 & 72.02 & 91.71 & 85.06 \\
 concatenation           & 85.39    & \textbf{11.98}    & \underline{89.44}            & 75.90 & \textbf{91.25} & \textbf{86.55} & \underline{95.67} & \underline{74.07} & 87.41 & 82.84 \\
  SCCSA       & \underline{85.57}   & 15.65                  & 89.33 & \textbf{79.36} & 88.66 & 85.65 & 95.13 & 73.55 & 90.97 & 81.94 \\
  Addition       & \underline{85.57}   & \underline{14.08}  & 88.87 & 74.47 & 88.71 & 85.36 & 95.40 & 73.57 & \textbf{93.02} & 85.13\\
 \midrule
 CSI(ours)   & \textbf{86.05}                                   & 15.82                                   & \textbf{89.49} & \underline{76.15} & 88.23 & \underline{86.26} & \textbf{95.87} & \textbf{74.14} & 91.31 & \underline{86.97} \\

\bottomrule
\end{tabular}
}
\label{skip}
\end{table*}

 To evaluate the superiority of the proposed CSI module in enhancing skip connections, we conduct a comparative study over several representative fusion strategies, including AG (Attention Gate in Attention U-Net (2018)), FFM (Feature Fusion Module in FusionU-Net (2024)), GAB (Group Aggregation Bridge in EGE-UNet (2023)), and SCCSA (Skip Connection Channel-Spatial Attention in BRAU-Net++ (2024)), as well as element-wise addition and channel concatenation.   In each case, only the CSI module in FIF-UNet is replaced with the alternative module to conduct the performance comparison.

 As shown in Table \ref{skip}, the proposed CSI module achieves the highest average Dice score of 86.05\%, outperforming all baselines with selected fusion modules. In terms of organ-wise performance, CSI achieves the best Dice scores on the aorta (89.49\%), liver (95.87\%), and pancreas (74.14\%), and ranks second-best on the gallbladder, right kidney, and stomach, showing strong generalization across organs with varying sizes and complexities.  Compared to the best baseline (SCCSA and Addition, both 85.57\%), the CSI enables the model to obtain a 0.48\% improvement, demonstrating the superiority of the fusion design.
In terms of HD95, Concatenation achieves the lowest value (11.98) due to its simple channel-stacking operation with fine-grained details, especially for small organs like kidneys, while it only obtains undesired overall performance due to limited semantic filtering.

 The AG module applies decoder features to reweight encoder features via gating mechanism, but the one-way gating strategy may discard valuable decoder information with reduced representation ability. The FFM fuses multi-scale encoder features without decoder features entirely, ignoring crucial context for semantic alignment. The GAB leverages decoder-generated masks to guide fusion of encoder features, in which, as decoding progresses, the resolution of these masks becomes coarser and noise-sensitive, to further diminish the model's effectiveness. 
The SCCSA has a similar structure to the proposed model, i.e., cascading channel and spatial attentions. However, the SCCSA first performs the addition operation for encoder and decoder features with a follow-up attention, which may limit its ability to model cross-level dependencies effectively. In contrast, the proposed CSI explicitly learns channel and spatial correlations between encoder and decoder features before performing element-wise addition, enabling more informative and meaningful semantic fusion.

 In summary, the results confirm that the CSI effectively facilitates multi-level semantic interaction between encoder and decoder,
enhancing the representation ability of skip connections and benefiting to superior multi-organ segmentation performance over existing fusion strategies.

\subsection{Generalization studies}
\begin{table*}[]
\caption{Generalization studies for the proposed modules}
\centering
\resizebox{1\textwidth}{!}{
\begin{tabular}{llc|lc|llllllll}
\toprule
\multirow{2}{*}{CoSE} & \multirow{2}{*}{CSI} & \multirow{2}{*}{MLF} & \multicolumn{2}{c|}{Average} & \multirow{2}{*}{Aorta} & \multirow{2}{*}{GB} & \multirow{2}{*}{KL} & \multirow{2}{*}{KR} & \multirow{2}{*}{Liver} & \multirow{2}{*}{PC} & \multirow{2}{*}{SP} & \multirow{2}{*}{SM} \\
            &      &   & DICE↑        & HD95↓        &                        &                     &                     &                     &                        &                     &                     &                     \\
\midrule
 $\times$   & $\times$  & $\times$     & 75.65  & 40.32 & 85.74 & 64.75 & 80.07 & 70.04 & 91.63 & 60.92 & 86.59 & 65.47  \\
\midrule
 \checkmark    & $\times$    & $\times$     & 77.91   & 36.62 & 86.24 & 65.13 & 78.44 & 75.44 & 94.11 & 59.06 & 86.88 & 77.95\\
 $\times$   & \checkmark    & $\times$    & 77.64   & 42.18 & 88.59 & 64.86 & 79.70 & 75.90 & 92.00 & 60.84 & 87.93 & 71.30\\
 $\times$   & $\times$    & \checkmark     & 77.85   & 29.66 & 87.61 & 71.11 & 82.89 & 70.60 & 94.12 & 52.77 & 88.39 & 75.28\\
\midrule
 \checkmark    & \checkmark    & $\times$     & 79.12   & 31.00 & 89.63 & 69.25 & 84.60 & 77.02 & 93.00 & 53.34 & 89.10 & 76.99\\
 \checkmark    & $\times$    & \checkmark     & 78.35   & 23.77 & 86.64 & 68.28 & 85.35 & 78.92 & 93.83 & 54.67 & 87.95 & 71.15\\
$\times$    & \checkmark    & \checkmark     & 78.45   & 40.91 & 87.94 & 66.66 & 82.96 & 78.37 & 93.20 & 57.89 & 85.00 & 75.58 \\
\midrule
 \checkmark    & \checkmark    & \checkmark     & 79.57   & 25.57                                   & 86.13 & 72.90 & 85.86 & 78.32 & 93.07 & 62.66 & 86.15 & 71.47 \\
\bottomrule
\end{tabular}
}
\label{generalization}
\end{table*}

To evaluate the generalization capability of the proposed modules, the original UNet is selected as the baseline to conduct the generalization experiments.
Similar to the ablation studies, the experimental design concerns the employment of the proposed technical modules separately or their combinations with the baseline to comprehensively evaluate the efficacy and applicability of the modules. 
As shown in Table \ref{generalization}, it can be seen that all the proposed technical modules provide desired performance improvements compared to the UNet baseline. 
Note that, instead of 70.11\%, our re-implementation result of the UNet is 75.65\% in Table \ref{generalization} due to the changed experimental configurations.
To be specific, the CoSE, CSI, and MLF modules improve the Dice score with 2.26\%, 1.99\% and 2.20\%, respectively.
The Dice score is improved by 3.47\%, 2.70\% and 2.80\% by employing every two modules in UNet.
Finally,  the UNet yields a performance improvement of 3.92\% by using all the proposed modules.

In summary, the above experimental results show that all the proposed modules contribute to performance improvement, even with different backbones, presenting expected generalization ability with consistent performance. Most importantly, we can also observe that the proposed modules have the ability to obtain higher performance improvements for simpler model architectures (i.e., original UNet).

\subsection{Qualitative results}
To further validate the quantitative findings and demonstrate the effectiveness of both the proposed modules and the overall FIF-UNet architecture, we provide comprehensive visualizations on the Synapse and ACDC.  For each dataset, the ablation visualizations are based on the models that achieved the highest Dice scores among the single-module and two-module combinations, 
\subsubsection{Experimental results on Synapse dataset}
\begin{figure*}[!htbp]
    \centering
    \includegraphics[width=1\textwidth]{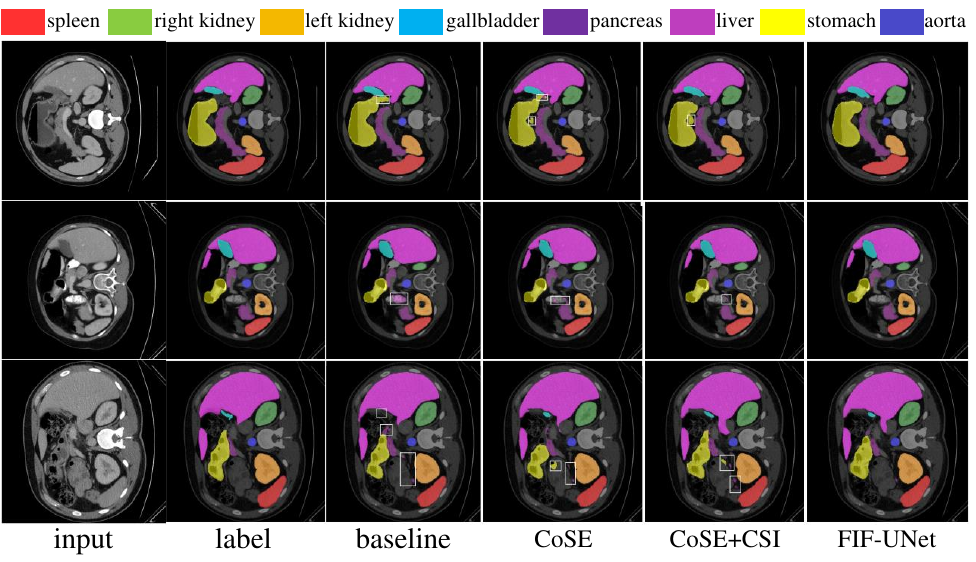}
    \caption{Visual comparison of ablation experiments on the Synapse dataset. “CoSE” represents the result of the model where the decoder block is the CoSE module; “CoSE+CSI” represents the result of adding the CSI module to the previous model; “FIF-UNet” represents the result of Small FIF-UNet. The part of the white rectangular box is the place where there is an obvious segmentation error.}
    \label{visiual}
\end{figure*}
\begin{figure*}[!htbp]
    \centering
    \includegraphics[width=1\textwidth]{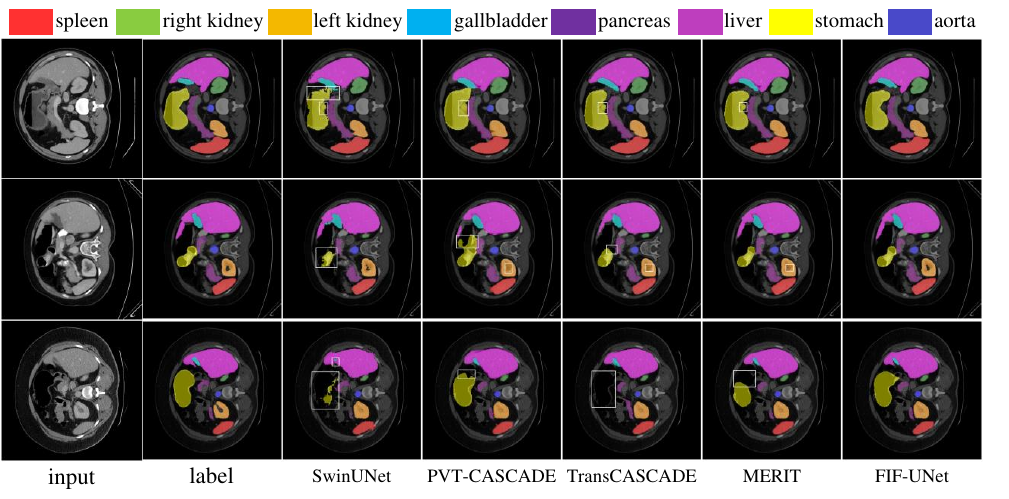}
    \caption{Visual comparison of different models on the Synapse dataset. “FIF-UNet” denotes Small FIF-UNet. “MERIT” denotes Cascaded MERIT. The part of the white rectangular box is the place where there is an obvious segmentation error.}
    \label{contrast}
\end{figure*}

Visualization results of ablation studies on the Synapse dataset are provided in Figure \ref{visiual}. 
Compared with the baseline model, the CoSE in the decoder considerably enhances the model to recognize foreground and background features, resulting in significantly reduced error regions. 
In particular, for the third row, the model successfully segments the gallbladder objects, while the baseline model fails to identify them.
With the combination of the CSI module, the approximate shape of each organ can be localized, suggesting that the model can accurately understand the structural characteristics of organs by obtaining informative features based on the different levels of semantic information between the encoder and the decoder. 
Compared with the previous model, the FIF-UNet has the ability to accurately capture the spatial distribution and morphological details of each organ by integrating semantic information of different scales, which helps reduce the misidentified scattering areas in the segmentation results.

In addition, the classical SwinUNet and three recent models (PVT-CASCADE, TransCASCADE, and Cascaded MERIT) are also selected to compare the visualization results with the proposed FIF-UNet.
As shown in Figure \ref{contrast}, compared with the selected models, the proposed FIF-UNet is capable of accurately locating all organs with precise details.
For the segmentation of the stomach, the selected baselines suffer from task challenges due to its high similarity with the background, i.e., can only partially identify the organ, or even fail to detect it. 
Fortunately, the proposed FIF-UNet model demonstrates significant advantages in accurately segmenting the entire stomach and greatly reducing the risk of misidentifying other regions as the stomach.
Furthermore, in the case of the second row, it can also be observed that compared to recent models, only the FIF-UNet can identify the hollow region in the middle of the left kidney, illustrating its advantages in capturing complex structural details.
From the above analysis, except for the quantitative metrics, more qualitative results also support the performance improvements over baselines.
\subsubsection{Experimental results on ACDC dataset}
\begin{figure*}[!htbp]
    \centering
    \includegraphics[width=1\textwidth]{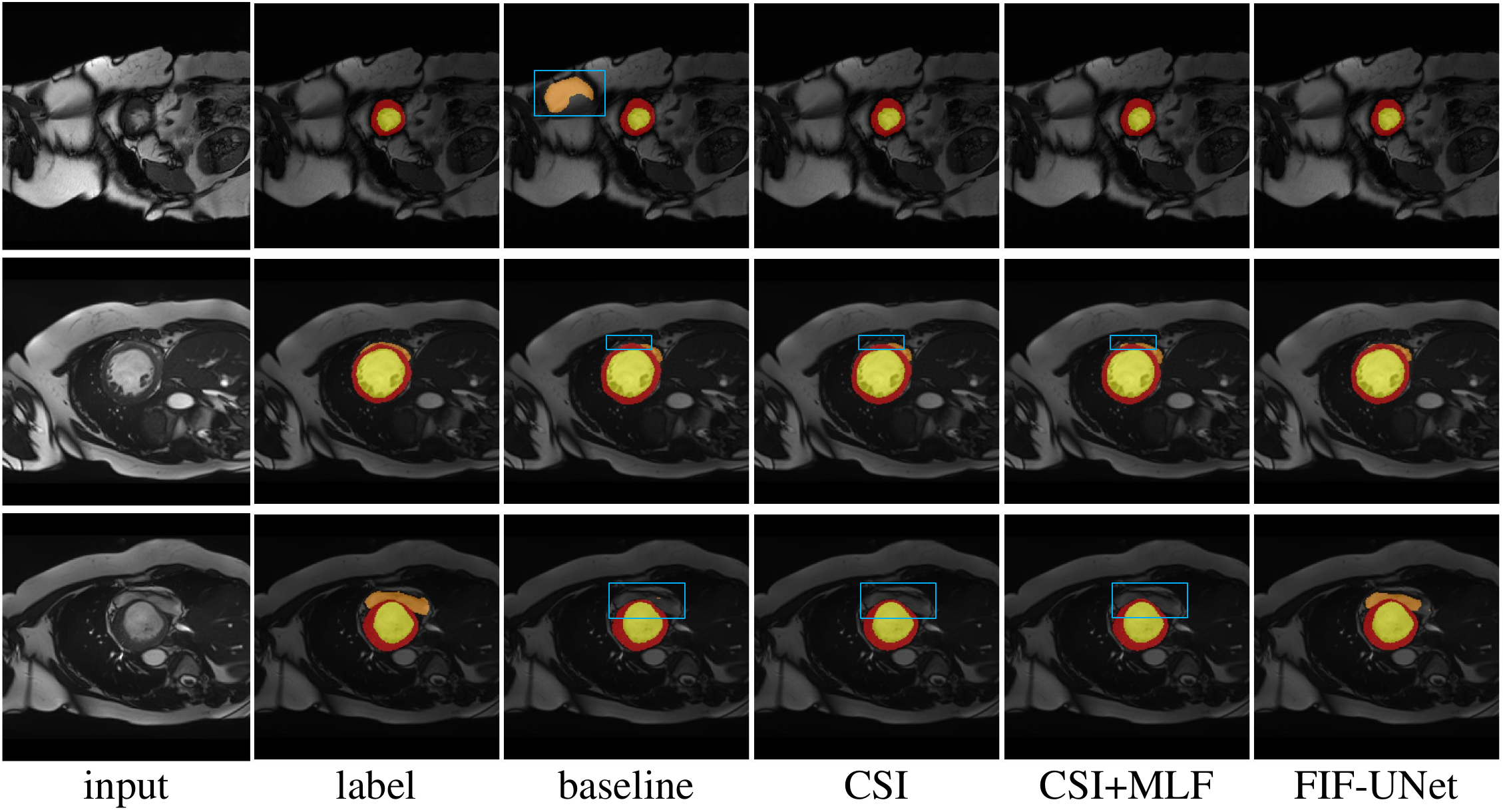}
    \caption{ Visual comparison of ablation experiments on the ACDC dataset. “CSI” represents the result of the model where the skip connection is the CSI module; “CSI+MLF” represents the result of adding the MLF module to the previous model; “FIF-UNet” represents the result of Small FIF-UNet. The part of the blue rectangular box is the place where there is an obvious segmentation error.}
    \label{acdcab}
\end{figure*}

\begin{figure*}[!htbp]
    \centering
    \includegraphics[width=1\textwidth]{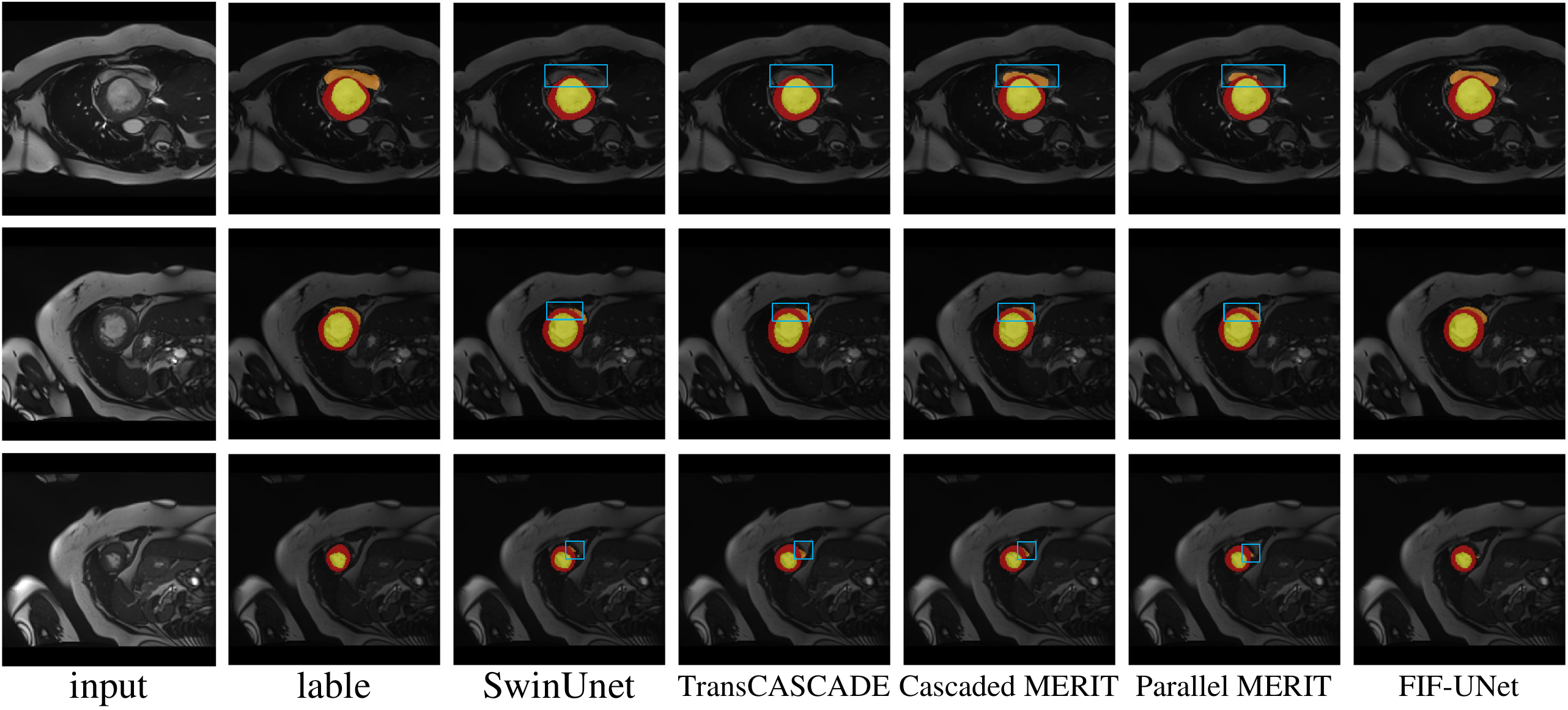}
    \caption{ Visual comparison of different models on the ACDC dataset. “FIF-UNet” denotes Small FIF-UNet. The part of the blue rectangular box is the place where there is an obvious segmentation error.}
    \label{acdccontr}
\end{figure*}

 The visualization results of the ablation experiments on the ACDC dataset are illustrated in Figure \ref{acdcab}. In general, the CSI module enhances the model’s ability to capture global semantic information to further improve background suppression and structure delineation.  As seen in the first row, the CSI can correct background misclassification from the baseline model to highlight its semantic enrichment. However, in ambiguous cases with high organ-background similarity, such as the RV shown in the third row, only the CSI fails to accurately localize the RV.
As in the second row, by adding the MLF module, the proposed model improves fine-grained detail through multi-scale feature fusion, i.e., the RV boundary is better recovered to indicate enhanced detail sensitivity. The full FIF-UNet model, incorporating CoSE, CSI and MLF, achieves the most robust segmentation results. The CoSE strengthens the model’s discriminative ability via channel-wise attention, enabling accurate identification of visually similar regions. In the third row, only FIF-UNet can successfully segment the RV, demonstrating the critical role of CoSE in refining feature representation and improving segmentation in challenging scenarios.

 Figure \ref{acdccontr} presents qualitative comparisons between FIF-UNet and four methods: SwinUNet, TransCASCADE, Cascaded MERIT, and Parallel MERIT. Similar to the results on the Synapse dataset, FIF-UNet consistently shows superior segmentation accuracy, particularly for organs with low contrast or indistinct boundaries. In the first row, SwinUNet and TransCASCADE fail to segment the RV, while Cascaded MERIT and Parallel MERIT only partially recover the structure. In contrast, FIF-UNet successfully identifies the entire RV, indicating its enhanced robustness in segmenting low-contrast, ambiguous structures. The second row features a highly blurred RV boundary, which provides considerable challenges to all comparative models. Even under these conditions, FIF-UNet accurately recovers more of the RV region than its counterparts, further demonstrating its strength in handling edge ambiguity. Finally, the third row shows that FIF-UNet produces the fewest false positives among all models, showing its superior performance and reduced susceptibility to misclassification.

\section{Conclusion}
In this work, we propose FIF-UNet, a novel U-shaped architecture designed to improve medical image segmentation by fully leveraging multi-level semantic features through effective interaction and fusion. The model integrates three plug-and-play modules: the Channel-Spatial Interaction (CSI) module, which enhances skip connections by modeling cross-stage dependencies from both channel and spatial perspectives; the Cascaded Convolution and Squeeze-and-Excitation (CoSE) module, which enhances the decoder’s ability to focus on critical regions by incorporating channel attention; and the Multi-Level Fusion (MLF) module, which aggregates decoder features across scales to preserve structural details. Extensive experiments on the Synapse and ACDC datasets demonstrate that FIF-UNet consistently outperforms state-of-the-art methods. Furthermore, ablation studies confirm the individual contributions of each module. Notably, all proposed modules are generalizable, making them applicable to other encoder-decoder segmentation frameworks.

In the future, we will attempt to apply the proposed model to other modalities of medical images to construct a generalized model for medical image segmentation.

\section*{Acknowledgements}
This work is supported by the National Natural Science Foundation of China under Grant U2333209, 62371323.

\bibliographystyle{cas-model2-names}

\bibliography{main}

\begin{thebibliography}{35}
\expandafter\ifx\csname natexlab\endcsname\relax\def\natexlab#1{#1}\fi
\providecommand{\url}[1]{\texttt{#1}}
\providecommand{\href}[2]{#2}
\providecommand{\path}[1]{#1}
\providecommand{\DOIprefix}{doi:}
\providecommand{\ArXivprefix}{arXiv:}
\providecommand{\URLprefix}{URL: }
\providecommand{\Pubmedprefix}{pmid:}
\providecommand{\doi}[1]{\href{http://dx.doi.org/#1}{\path{#1}}}
\providecommand{\Pubmed}[1]{\href{pmid:#1}{\path{#1}}}
\providecommand{\bibinfo}[2]{#2}
\ifx\xfnm\relax \def\xfnm[#1]{\unskip,\space#1}\fi
\bibitem[{Azad et~al.(2022a)Azad, Aghdam, Rauland, Jia, Avval, Bozorgpour, Karimijafarbigloo, Cohen, Adeli and Merhof}]{Azadetal2022}
\bibinfo{author}{Azad, R.}, \bibinfo{author}{Aghdam, E.K.}, \bibinfo{author}{Rauland, A.}, \bibinfo{author}{Jia, Y.}, \bibinfo{author}{Avval, A.H.}, \bibinfo{author}{Bozorgpour, A.}, \bibinfo{author}{Karimijafarbigloo, S.}, \bibinfo{author}{Cohen, J.P.}, \bibinfo{author}{Adeli, E.}, \bibinfo{author}{Merhof, D.}, \bibinfo{year}{2022}a.
\newblock \bibinfo{title}{Medical image segmentation review: The success of u-net}.
\newblock \bibinfo{journal}{arXiv:2211.14830} .
\bibitem[{Azad et~al.(2022b)Azad, Heidari, Wu and Merhof}]{Azadetal2022a}
\bibinfo{author}{Azad, R.}, \bibinfo{author}{Heidari, M.}, \bibinfo{author}{Wu, Y.}, \bibinfo{author}{Merhof, D.}, \bibinfo{year}{2022}b.
\newblock \bibinfo{title}{Contextual attention network: Transformer meets u-net}, in: \bibinfo{booktitle}{International Workshop on Machine Learning in Medical Imaging}, pp. \bibinfo{pages}{377--386}.
\bibitem[{Bernard et~al.(2018)Bernard, Lalande, Zotti, Cervenansky, Yang, Heng, Cetin, Lekadir, Camara, Ballester and others.}]{Bernardetal2018}
\bibinfo{author}{Bernard, O.}, \bibinfo{author}{Lalande, A.}, \bibinfo{author}{Zotti, C.}, \bibinfo{author}{Cervenansky, F.}, \bibinfo{author}{Yang, X.}, \bibinfo{author}{Heng, P.A.}, \bibinfo{author}{Cetin, I.}, \bibinfo{author}{Lekadir, K.}, \bibinfo{author}{Camara, O.}, \bibinfo{author}{Ballester, M.A.G.}, \bibinfo{author}{others.}, \bibinfo{year}{2018}.
\newblock \bibinfo{title}{Deep learning techniques for automatic mri cardiac multi-structures segmentation and diagnosis: is the problem solved?}
\newblock \bibinfo{journal}{IEEE transactions on medical imaging} \bibinfo{volume}{37}, \bibinfo{pages}{2514--2525}.
\bibitem[{Cao et~al.(2022)Cao, Wang, Chen, Jiang, Zhang, Tian and Wang}]{Caoetal2022}
\bibinfo{author}{Cao, H.}, \bibinfo{author}{Wang, Y.}, \bibinfo{author}{Chen, J.}, \bibinfo{author}{Jiang, D.}, \bibinfo{author}{Zhang, X.}, \bibinfo{author}{Tian, Q.}, \bibinfo{author}{Wang, M.}, \bibinfo{year}{2022}.
\newblock \bibinfo{title}{Swin-unet: Unet-like pure transformer for medical image segmentation}, in: \bibinfo{booktitle}{European conference on computer vision}, pp. \bibinfo{pages}{205--218}.
\bibitem[{Chen et~al.(2021)Chen, Lu, Yu, Luo, Adeli, Wang, Lu, Yuille and Zhou}]{Chenetal2021}
\bibinfo{author}{Chen, J.}, \bibinfo{author}{Lu, Y.}, \bibinfo{author}{Yu, Q.}, \bibinfo{author}{Luo, X.}, \bibinfo{author}{Adeli, E.}, \bibinfo{author}{Wang, Y.}, \bibinfo{author}{Lu, L.}, \bibinfo{author}{Yuille, A.L.}, \bibinfo{author}{Zhou, Y.}, \bibinfo{year}{2021}.
\newblock \bibinfo{title}{Transunet: Transformers make strong encoders for medical image segmentation}.
\newblock \bibinfo{journal}{arXiv:2102.04306} .
\bibitem[{Dong et~al.(2021)Dong, Wang, Fan, Li, Fu and Shao}]{Dongetal2021}
\bibinfo{author}{Dong, B.}, \bibinfo{author}{Wang, W.}, \bibinfo{author}{Fan, D.P.}, \bibinfo{author}{Li, J.}, \bibinfo{author}{Fu, H.}, \bibinfo{author}{Shao, L.}, \bibinfo{year}{2021}.
\newblock \bibinfo{title}{Polyp-pvt: Polyp segmentation with pyramid vision transformers}.
\newblock \bibinfo{journal}{arXiv:2108.06932} .
\bibitem[{Dosovitskiy et~al.(2020)Dosovitskiy, Beyer, Kolesnikov, Weissenborn, Zhai, Unterthiner, Dehghani, Minderer, Heigold, Gelly and others.}]{Dosovitskiyetal2020}
\bibinfo{author}{Dosovitskiy, A.}, \bibinfo{author}{Beyer, L.}, \bibinfo{author}{Kolesnikov, A.}, \bibinfo{author}{Weissenborn, D.}, \bibinfo{author}{Zhai, X.}, \bibinfo{author}{Unterthiner, T.}, \bibinfo{author}{Dehghani, M.}, \bibinfo{author}{Minderer, M.}, \bibinfo{author}{Heigold, G.}, \bibinfo{author}{Gelly, S.}, \bibinfo{author}{others.}, \bibinfo{year}{2020}.
\newblock \bibinfo{title}{An image is worth 16x16 words: Transformers for image recognition at scale}.
\newblock \bibinfo{journal}{arXiv:2010.11929} .
\bibitem[{Hatamizadeh et~al.(2022)Hatamizadeh, Tang, Nath, Yang, Myronenko, Landman, Roth and Xu}]{Hatamizadehetal2022}
\bibinfo{author}{Hatamizadeh, A.}, \bibinfo{author}{Tang, Y.}, \bibinfo{author}{Nath, V.}, \bibinfo{author}{Yang, D.}, \bibinfo{author}{Myronenko, A.}, \bibinfo{author}{Landman, B.}, \bibinfo{author}{Roth, H.R.}, \bibinfo{author}{Xu, D.}, \bibinfo{year}{2022}.
\newblock \bibinfo{title}{Unetr: Transformers for 3d medical image segmentation}, in: \bibinfo{booktitle}{Proceedings of the IEEE/CVF winter conference on applications of computer vision}, pp. \bibinfo{pages}{574--584}.
\bibitem[{Heidari et~al.(2023)Heidari, Kazerouni, Soltany, Azad, Aghdam, Cohen-Adad and Merhof}]{Heidarietal2023}
\bibinfo{author}{Heidari, M.}, \bibinfo{author}{Kazerouni, A.}, \bibinfo{author}{Soltany, M.}, \bibinfo{author}{Azad, R.}, \bibinfo{author}{Aghdam, E.K.}, \bibinfo{author}{Cohen-Adad, J.}, \bibinfo{author}{Merhof, D.}, \bibinfo{year}{2023}.
\newblock \bibinfo{title}{Hiformer: Hierarchical multi-scale representations using transformers for medical image segmentation}, in: \bibinfo{booktitle}{Proceedings of the IEEE/CVF winter conference on applications of computer vision}, pp. \bibinfo{pages}{6202--6212}.
\bibitem[{Hu et~al.(2018)Hu, Shen and Sun}]{Huetal2018}
\bibinfo{author}{Hu, J.}, \bibinfo{author}{Shen, L.}, \bibinfo{author}{Sun, G.}, \bibinfo{year}{2018}.
\newblock \bibinfo{title}{Squeeze-and-excitation networks}, in: \bibinfo{booktitle}{Proceedings of the IEEE conference on computer vision and pattern recognition}, pp. \bibinfo{pages}{7132--7141}.
\bibitem[{Huang et~al.(2021)Huang, Deng, Li and Yuan}]{Huangetal2021}
\bibinfo{author}{Huang, X.}, \bibinfo{author}{Deng, Z.}, \bibinfo{author}{Li, D.}, \bibinfo{author}{Yuan, X.}, \bibinfo{year}{2021}.
\newblock \bibinfo{title}{Missformer: An effective medical image segmentation transformer}.
\newblock \bibinfo{journal}{arXiv:2109.07162} .
\bibitem[{Isensee et~al.(2021)Isensee, Jaeger, Kohl, Petersen and Maier-Hein}]{isensee2021nnu}
\bibinfo{author}{Isensee, F.}, \bibinfo{author}{Jaeger, P.F.}, \bibinfo{author}{Kohl, S.A.}, \bibinfo{author}{Petersen, J.}, \bibinfo{author}{Maier-Hein, K.H.}, \bibinfo{year}{2021}.
\newblock \bibinfo{title}{nnu-net: a self-configuring method for deep learning-based biomedical image segmentation}.
\newblock \bibinfo{journal}{Nature methods} \bibinfo{volume}{18}, \bibinfo{pages}{203--211}.
\bibitem[{Jin et~al.(2019)Jin, Meng, Pham, Chen, Wei and Su}]{Jinetal2019}
\bibinfo{author}{Jin, Q.}, \bibinfo{author}{Meng, Z.}, \bibinfo{author}{Pham, T.D.}, \bibinfo{author}{Chen, Q.}, \bibinfo{author}{Wei, L.}, \bibinfo{author}{Su, R.}, \bibinfo{year}{2019}.
\newblock \bibinfo{title}{Dunet: A deformable network for retinal vessel segmentation}.
\newblock \bibinfo{journal}{Knowledge-Based Systems} \bibinfo{volume}{178}, \bibinfo{pages}{149--162}.
\bibitem[{Lan et~al.(2024)Lan, Cai, Jiang, Liu, Li and Zhang}]{lan2024brau}
\bibinfo{author}{Lan, L.}, \bibinfo{author}{Cai, P.}, \bibinfo{author}{Jiang, L.}, \bibinfo{author}{Liu, X.}, \bibinfo{author}{Li, Y.}, \bibinfo{author}{Zhang, Y.}, \bibinfo{year}{2024}.
\newblock \bibinfo{title}{Brau-net++: U-shaped hybrid cnn-transformer network for medical image segmentation}.
\newblock \bibinfo{journal}{arXiv preprint arXiv:2401.00722} .
\bibitem[{Landman et~al.(2015)Landman, Xu, Igelsias, Styner, Langerak and Klein}]{Landmanetal2015}
\bibinfo{author}{Landman, B.}, \bibinfo{author}{Xu, Z.}, \bibinfo{author}{Igelsias, J.}, \bibinfo{author}{Styner, M.}, \bibinfo{author}{Langerak, T.}, \bibinfo{author}{Klein, A.}, \bibinfo{year}{2015}.
\newblock \bibinfo{title}{Miccai multi-atlas labeling beyond the cranial vault--workshop and challenge}, in: \bibinfo{booktitle}{Proc. MICCAI Multi-Atlas Labeling Beyond Cranial Vault—Workshop Challenge}, p.~\bibinfo{pages}{12}.
\bibitem[{Li et~al.(2024)Li, Lyu and Wang}]{Lieatl2024}
\bibinfo{author}{Li, Z.}, \bibinfo{author}{Lyu, H.}, \bibinfo{author}{Wang, J.}, \bibinfo{year}{2024}.
\newblock \bibinfo{title}{Fusionu-net: U-net with enhanced skip connection for pathology image segmentation}, in: \bibinfo{booktitle}{Asian Conference on Machine Learning}, pp. \bibinfo{pages}{694--706}.
\bibitem[{Liu et~al.(2022)Liu, Mao, Wu, Feichtenhofer, Darrell and Xie}]{Liuetal2022}
\bibinfo{author}{Liu, Z.}, \bibinfo{author}{Mao, H.}, \bibinfo{author}{Wu, C.Y.}, \bibinfo{author}{Feichtenhofer, C.}, \bibinfo{author}{Darrell, T.}, \bibinfo{author}{Xie, S.}, \bibinfo{year}{2022}.
\newblock \bibinfo{title}{A convnet for the 2020s}, in: \bibinfo{booktitle}{Proceedings of the IEEE/CVF conference on computer vision and pattern recognition}, pp. \bibinfo{pages}{11976--11986}.
\bibitem[{Loshchilov and Hutter(2017)}]{LoshchilovHutter2017}
\bibinfo{author}{Loshchilov, I.}, \bibinfo{author}{Hutter, F.}, \bibinfo{year}{2017}.
\newblock \bibinfo{title}{Decoupled weight decay regularization}.
\newblock \bibinfo{journal}{arXiv:1711.05101} .
\bibitem[{Oktay et~al.(2018)Oktay, Schlemper, Folgoc, Lee, Heinrich, Misawa, Mori, McDonagh, Hammerla, Kainz and others.}]{Oktayetal2018}
\bibinfo{author}{Oktay, O.}, \bibinfo{author}{Schlemper, J.}, \bibinfo{author}{Folgoc, L.L.}, \bibinfo{author}{Lee, M.}, \bibinfo{author}{Heinrich, M.}, \bibinfo{author}{Misawa, K.}, \bibinfo{author}{Mori, K.}, \bibinfo{author}{McDonagh, S.}, \bibinfo{author}{Hammerla, N.Y.}, \bibinfo{author}{Kainz, B.}, \bibinfo{author}{others.}, \bibinfo{year}{2018}.
\newblock \bibinfo{title}{Attention u-net: Learning where to look for the pancreas}.
\newblock \bibinfo{journal}{arXiv:1804.03999} .
\bibitem[{Punn and Agarwal(2020)}]{punnAgarwal2020}
\bibinfo{author}{Punn, N.S.}, \bibinfo{author}{Agarwal, S.}, \bibinfo{year}{2020}.
\newblock \bibinfo{title}{Inception u-net architecture for semantic segmentation to identify nuclei in microscopy cell images}.
\newblock \bibinfo{journal}{ACM Transactions on Multimedia Computing, Communications, and Applications (TOMM)} \bibinfo{volume}{16}, \bibinfo{pages}{1--15}.
\bibitem[{Rahman and Marculescu(2023)}]{RahmanMarculescu2023}
\bibinfo{author}{Rahman, M.M.}, \bibinfo{author}{Marculescu, R.}, \bibinfo{year}{2023}.
\newblock \bibinfo{title}{Medical image segmentation via cascaded attention decoding}, in: \bibinfo{booktitle}{Proceedings of the IEEE/CVF Winter Conference on Applications of Computer Vision}, pp. \bibinfo{pages}{6222--6231}.
\bibitem[{Rahman and Marculescu(2024)}]{RahmanMarculescu2024}
\bibinfo{author}{Rahman, M.M.}, \bibinfo{author}{Marculescu, R.}, \bibinfo{year}{2024}.
\newblock \bibinfo{title}{Multi-scale hierarchical vision transformer with cascaded attention decoding for medical image segmentation}, in: \bibinfo{booktitle}{Medical Imaging with Deep Learning}, pp. \bibinfo{pages}{1526--1544}.
\bibitem[{Ronneberger et~al.(2015)Ronneberger, Fischer and Brox}]{Ronnebergeretal2015}
\bibinfo{author}{Ronneberger, O.}, \bibinfo{author}{Fischer, P.}, \bibinfo{author}{Brox, T.}, \bibinfo{year}{2015}.
\newblock \bibinfo{title}{U-net: Convolutional networks for biomedical image segmentation}, in: \bibinfo{booktitle}{Medical image computing and computer-assisted intervention--MICCAI 2015: 18th international conference, Munich, Germany, October 5-9, 2015, proceedings, part III 18}, pp. \bibinfo{pages}{234--241}.
\bibitem[{Ruan et~al.(2023)Ruan, Xie, Gao, Liu and Fu}]{ruanetal2023}
\bibinfo{author}{Ruan, J.}, \bibinfo{author}{Xie, M.}, \bibinfo{author}{Gao, J.}, \bibinfo{author}{Liu, T.}, \bibinfo{author}{Fu, Y.}, \bibinfo{year}{2023}.
\newblock \bibinfo{title}{Ege-unet: an efficient group enhanced unet for skin lesion segmentation}, in: \bibinfo{booktitle}{International conference on medical image computing and computer-assisted intervention}, pp. \bibinfo{pages}{481--490}.
\bibitem[{Tan and Le(2019)}]{TanLe2019}
\bibinfo{author}{Tan, M.}, \bibinfo{author}{Le, Q.}, \bibinfo{year}{2019}.
\newblock \bibinfo{title}{Efficientnet: Rethinking model scaling for convolutional neural networks}, in: \bibinfo{booktitle}{International conference on machine learning}, pp. \bibinfo{pages}{6105--6114}.
\bibitem[{Tragakis et~al.(2023)Tragakis, Kaul, Murray-Smith and Husmeier}]{tragakisetal2023}
\bibinfo{author}{Tragakis, A.}, \bibinfo{author}{Kaul, C.}, \bibinfo{author}{Murray-Smith, R.}, \bibinfo{author}{Husmeier, D.}, \bibinfo{year}{2023}.
\newblock \bibinfo{title}{The fully convolutional transformer for medical image segmentation}, in: \bibinfo{booktitle}{Proceedings of the IEEE/CVF Winter Conference on Applications of Computer Vision}, pp. \bibinfo{pages}{3660--3669}.
\bibitem[{Tu et~al.(2022)Tu, Talebi, Zhang, Yang, Milanfar, Bovik and Li}]{Tuetal2022}
\bibinfo{author}{Tu, Z.}, \bibinfo{author}{Talebi, H.}, \bibinfo{author}{Zhang, H.}, \bibinfo{author}{Yang, F.}, \bibinfo{author}{Milanfar, P.}, \bibinfo{author}{Bovik, A.}, \bibinfo{author}{Li, Y.}, \bibinfo{year}{2022}.
\newblock \bibinfo{title}{Maxvit: Multi-axis vision transformer}, in: \bibinfo{booktitle}{European conference on computer vision}, pp. \bibinfo{pages}{459--479}.
\bibitem[{Valanarasu et~al.(2021)Valanarasu, Oza, Hacihaliloglu and Patel}]{Valanarasuetal2021}
\bibinfo{author}{Valanarasu, J.M.J.}, \bibinfo{author}{Oza, P.}, \bibinfo{author}{Hacihaliloglu, I.}, \bibinfo{author}{Patel, V.M.}, \bibinfo{year}{2021}.
\newblock \bibinfo{title}{Medical transformer: Gated axial-attention for medical image segmentation}, in: \bibinfo{booktitle}{Medical image computing and computer assisted intervention--MICCAI 2021: 24th international conference, Strasbourg, France, September 27--October 1, 2021, proceedings, part I 24}, pp. \bibinfo{pages}{36--46}.
\bibitem[{Vaswani et~al.(2017)Vaswani, Shazeer, Parmar, Uszkoreit, Jones, Gomez, Kaiser and Polosukhin}]{Vaswanietal2017}
\bibinfo{author}{Vaswani, A.}, \bibinfo{author}{Shazeer, N.}, \bibinfo{author}{Parmar, N.}, \bibinfo{author}{Uszkoreit, J.}, \bibinfo{author}{Jones, L.}, \bibinfo{author}{Gomez, A.N.}, \bibinfo{author}{Kaiser, {\L}.}, \bibinfo{author}{Polosukhin, I.}, \bibinfo{year}{2017}.
\newblock \bibinfo{title}{Attention is all you need}.
\newblock \bibinfo{journal}{Advances in neural information processing systems} \bibinfo{volume}{30}.
\bibitem[{Wang et~al.(2022)Wang, Huang, Tang, Meng, Su and Song}]{Wangetal2022}
\bibinfo{author}{Wang, J.}, \bibinfo{author}{Huang, Q.}, \bibinfo{author}{Tang, F.}, \bibinfo{author}{Meng, J.}, \bibinfo{author}{Su, J.}, \bibinfo{author}{Song, S.}, \bibinfo{year}{2022}.
\newblock \bibinfo{title}{Stepwise feature fusion: Local guides global}, in: \bibinfo{booktitle}{International Conference on Medical Image Computing and Computer-Assisted Intervention}, pp. \bibinfo{pages}{110--120}.
\bibitem[{Wang et~al.(2021)Wang, Chen, Ding, Yu, Zha and Li}]{Wangetal2021}
\bibinfo{author}{Wang, W.}, \bibinfo{author}{Chen, C.}, \bibinfo{author}{Ding, M.}, \bibinfo{author}{Yu, H.}, \bibinfo{author}{Zha, S.}, \bibinfo{author}{Li, J.}, \bibinfo{year}{2021}.
\newblock \bibinfo{title}{Transbts: Multimodal brain tumor segmentation using transformer}, in: \bibinfo{booktitle}{International Conference on Medical Image Computing and Computer-Assisted Intervention, Springer}, pp. \bibinfo{pages}{109--119}.
\bibitem[{Wu et~al.(2023)Wu, Liao, Chen, Wang, Chen, Gao and Wu}]{Wuetal2023}
\bibinfo{author}{Wu, Y.}, \bibinfo{author}{Liao, K.}, \bibinfo{author}{Chen, J.}, \bibinfo{author}{Wang, J.}, \bibinfo{author}{Chen, D.Z.}, \bibinfo{author}{Gao, H.}, \bibinfo{author}{Wu, J.}, \bibinfo{year}{2023}.
\newblock \bibinfo{title}{D-former: A u-shaped dilated transformer for 3d medical image segmentation}.
\newblock \bibinfo{journal}{Neural Computing and Applications} \bibinfo{volume}{35}, \bibinfo{pages}{1931--1944}.
\bibitem[{You et~al.(2022)You, Zhao, Liu, Dong, Chinchali, Topcu, Staib and Duncan}]{Youetal2022}
\bibinfo{author}{You, C.}, \bibinfo{author}{Zhao, R.}, \bibinfo{author}{Liu, F.}, \bibinfo{author}{Dong, S.}, \bibinfo{author}{Chinchali, S.}, \bibinfo{author}{Topcu, U.}, \bibinfo{author}{Staib, L.}, \bibinfo{author}{Duncan, J.}, \bibinfo{year}{2022}.
\newblock \bibinfo{title}{Class-aware adversarial transformers for medical image segmentation}.
\newblock \bibinfo{journal}{Advances in Neural Information Processing Systems} \bibinfo{volume}{35}, \bibinfo{pages}{29582--29596}.
\bibitem[{Zhou et~al.(2021)Zhou, Kang, Jin, Yang, Lian, Jiang, Hou and Feng}]{Zhouetal2021}
\bibinfo{author}{Zhou, D.}, \bibinfo{author}{Kang, B.}, \bibinfo{author}{Jin, X.}, \bibinfo{author}{Yang, L.}, \bibinfo{author}{Lian, X.}, \bibinfo{author}{Jiang, Z.}, \bibinfo{author}{Hou, Q.}, \bibinfo{author}{Feng, J.}, \bibinfo{year}{2021}.
\newblock \bibinfo{title}{Deepvit: Towards deeper vision transformer}.
\newblock \bibinfo{journal}{arXiv:2103.11886} .
\bibitem[{Zhou et~al.(2019)Zhou, Siddiquee, Tajbakhsh and Liang}]{Zhouetal2019}
\bibinfo{author}{Zhou, Z.}, \bibinfo{author}{Siddiquee, M.M.R.}, \bibinfo{author}{Tajbakhsh, N.}, \bibinfo{author}{Liang, J.}, \bibinfo{year}{2019}.
\newblock \bibinfo{title}{Unet++: Redesigning skip connections to exploit multiscale features in image segmentation}.
\newblock \bibinfo{journal}{IEEE transactions on medical imaging} \bibinfo{volume}{39}, \bibinfo{pages}{1856--1867}.

\end{thebibliography}

\end{document}